\documentclass[10pt,twocolumn,letterpaper]{article}

\usepackage{cvpr}              

\usepackage{float}
\usepackage{bigstrut}
\usepackage{comment}
\usepackage{booktabs}
\usepackage{multirow}
\usepackage{url}
\usepackage{soul}
\usepackage{fontawesome}
\usepackage{subcaption}
\usepackage[font={footnotesize}]{caption}
\usepackage{cuted}
\usepackage{bm}  
\usepackage{amsmath}  
\usepackage{amssymb}  
\usepackage{lineno}

\usepackage[table,xcdraw,dvipsnames]{xcolor}
\definecolor{tabfirst}{rgb}{1, 0.7, 0.7} 
\definecolor{tabsecond}{rgb}{1, 0.85, 0.7} 
\newcommand{\tsb}[1]{\textsubscript{\textcolor{gray}{$\pm$#1}}}

{\end{list}}
\usepackage{amsmath}

\expandafter\def\expandafter\normalsize\expandafter{%
    \normalsize%
    \setlength\abovedisplayskip{0pt}%
    \setlength\belowdisplayskip{0pt}%
    \setlength\abovedisplayshortskip{3pt}%
    \setlength\belowdisplayshortskip{0pt}%
}

\definecolor{cvprblue}{rgb}{0.21,0.49,0.74}
\usepackage[pagebackref,breaklinks,colorlinks,citecolor=cvprblue]{hyperref}
\usepackage{siunitx}
\usepackage{diagbox}
\def\paperID{70} 
\def\confName{3DV\xspace}
\def\confYear{2026\xspace}

\title{GaussianArt: Unified Modeling of Geometry and Motion for Articulated Objects}

\author{
Licheng Shen\textsuperscript{1,2*},
Saining Zhang\textsuperscript{1,2,3*$\dagger$}, 
Honghan Li\textsuperscript{1,2,3*}, 
Peilin Yang\textsuperscript{1,4},\\
Zihao Huang\textsuperscript{1,5}, 
Zongzheng Zhang\textsuperscript{1,2}, 
Hao Zhao\textsuperscript{2,1$\ddagger$}
\\
\\
\textsuperscript{1}BAAI \quad
\textsuperscript{2}AIR, THU \quad 
\textsuperscript{3}NTU \quad
\textsuperscript{4}BIT \quad
\textsuperscript{5}HUST \quad
}

\begin{document}
\maketitle
{\let\thefootnote\relax
 \footnotetext{$^*$ Equal contribution.}
 \footnotetext{$^\dagger$ Project leader.}
 \footnotetext{$^\ddagger$ Corresponding author.}
}

\begin{strip}
    \centering
    \includegraphics[width=0.95\linewidth]{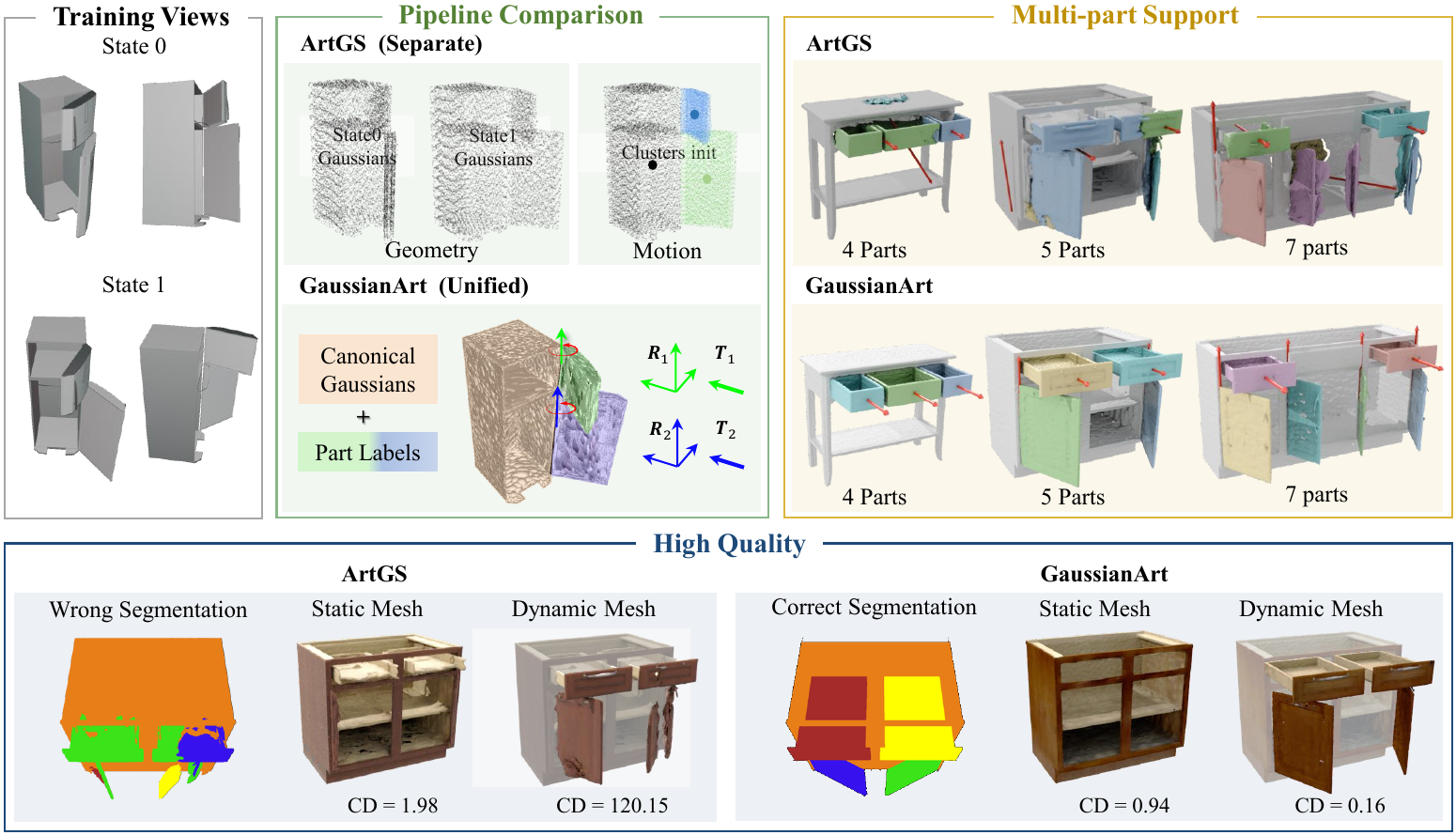}
    \vspace{-2mm}
    \captionof{figure}{ArtGS adopts a separate pipeline, while GaussianArt introduces a unified representation that jointly models geometry and motion. Compared to ArtGS, which struggles with wrong part assignments and axis errors beyond 2–3 parts, our method robustly handles complex objects with many parts. In terms of quality, ArtGS produces wrong segmentation and high errors (CD = 1.98 / 120.15 for static/dynamic), while GaussianArt achieves correct segmentation and much lower errors (CD = 0.94 / 0.16). These results highlight the scalability and accuracy benefits of our unified design.}
    \vspace{-3mm}
    \label{fig:teaser}
\end{strip}

\begin{abstract}
Reconstructing articulated objects is essential for building digital twins of interactive environments. However, prior methods typically decouple geometry and motion by first reconstructing object shape in distinct states and then estimating articulation through post-hoc alignment. This separation complicates the reconstruction pipeline and restricts scalability, especially for objects with complex, multi-part articulation. We introduce a \textbf{unified} representation that jointly models geometry and motion using articulated 3D Gaussians. This formulation improves robustness in motion decomposition and supports articulated objects with up to 20 parts, significantly outperforming prior approaches that often struggle beyond 2–3 parts due to brittle initialization. To systematically assess scalability and generalization, we propose MPArt-90, a new benchmark consisting of 90 articulated objects across 20 categories, each with diverse part counts and motion configurations. Extensive experiments show that our method consistently achieves superior accuracy in part-level geometry reconstruction and motion estimation across a broad range of object types. We further demonstrate applicability to downstream tasks such as robotic simulation and human-scene interaction modeling, highlighting the potential of unified articulated representations in scalable physical modeling.
\href{https://sainingzhang.github.io/project/gaussianart/}{Project Page.}
\end{abstract}
    
\section{Introduction}
Reconstructing articulated objects plays a central role in creating digital twins for robotic simulation and generic interaction modeling \cite{liu2024survey, pejic2022articulated, hu2018functionality}. While recent progress \cite{Jiang_2022_ditto,liu2023paris,weng2024digitaltwin,swaminathan24leia,guo2025articulatedgs,liu2025artgs,lin2025splart,yu2025part} has been made, most existing pipelines \cite{weng2024digitaltwin,liu2025artgs,lin2025splart,yu2025part} adopt a decoupled design as shown in \cref{fig:teaser}: they first reconstruct geometry from two static observations and subsequently infer motion through part-wise alignment. This separation not only introduces redundant modeling and brittle optimization, \textbf{but more critically}, breaks the physical consistency across object states—the same region of geometry is reconstructed independently per state without a coherent articulation structure.

As shown in \cref{fig:teaser} (top-left and top-middle), ArtGS \cite{liu2025artgs} exemplifies this limitation: it separately models geometry and motion using per-state 3D Gaussians, then relies on clustering to estimate part motion. Such pipelines often suffer from incorrect part grouping and axis misalignment, especially when handling more than 2–3 moving parts.

Several recent systems, including DigitalTwinArt \cite{weng2024digitaltwin} and ArtGS \cite{liu2025artgs}, attempt to address multi-part articulation by leveraging learned part-field or clustering heuristics. However, they are constrained by: (1) separate optimization of geometry and motion; (2) brittle initialization, often failing to produce reliable part decomposition under complex configurations; and (3) limited scalability, rarely tested beyond a dozen objects, and often constrained to two-part motion. Moreover, current benchmarks are narrow in scope—most evaluate on fewer than 20 objects with constrained topology and articulation patterns.

We introduce \textbf{GaussianArt}, a physically consistent and scalable framework for articulated object reconstruction. Unlike prior methods, GaussianArt adopts a unified representation based on articulated 3D Gaussian primitives, where each Gaussian simultaneously encodes its part affiliation (via learned soft assignments) and its rigid motion (as a mixture of motion bases). This formulation enables geometry and motion to be co-optimized within a single differentiable structure, ensuring cross-state consistency and interpretability, as visualized in \cref{fig:teaser} (middle row).

To evaluate scalability, we construct \textbf{MPArt-90}, a benchmark of 90 articulated objects across 20 categories, each with observations in two states and full articulation ground truth. As shown in \cref{fig:teaser} (top-right), GaussianArt correctly recovers fine-grained part structure and motion parameters, even with 7 components. In contrast, ArtGS suffers from cluster collapse, resulting in incorrect part grouping and misaligned joints. In terms of quality, \cref{fig:teaser} (bottom row) highlights the significant gains in both segmentation accuracy and reconstruction fidelity. GaussianArt reduces the dynamic part Chamfer Distance (CD) from 120.15 to 0.16, demonstrating an order-of-magnitude improvement.

In summary, our contributions are:

\begin{itemize}
    \item We propose GaussianArt, a reconstruction method that jointly models geometry and motion using articulated 3D Gaussians, enabling consistent reasoning across states.

    \item We design a soft-to-hard training paradigm that progressively refines part segmentation and rigid motion parameters, improving robustness on complex multi-part objects.
    
    \item We evaluate on MPArt-90, the largest benchmark to date for articulated object reconstruction, featuring 90 objects across 20 categories, with up to 20 parts (19 movable parts) and ground-truth motion annotations.

    \item Our method significantly outperforms ArtGS in both geometry and motion accuracy, and supports deployment in downstream tasks such as robotic manipulation and human-scene interaction (HSI) modeling.
\end{itemize}	

\label{sec:method}
\begin{figure*}[t] 
\centering
\includegraphics[width=0.95\linewidth]{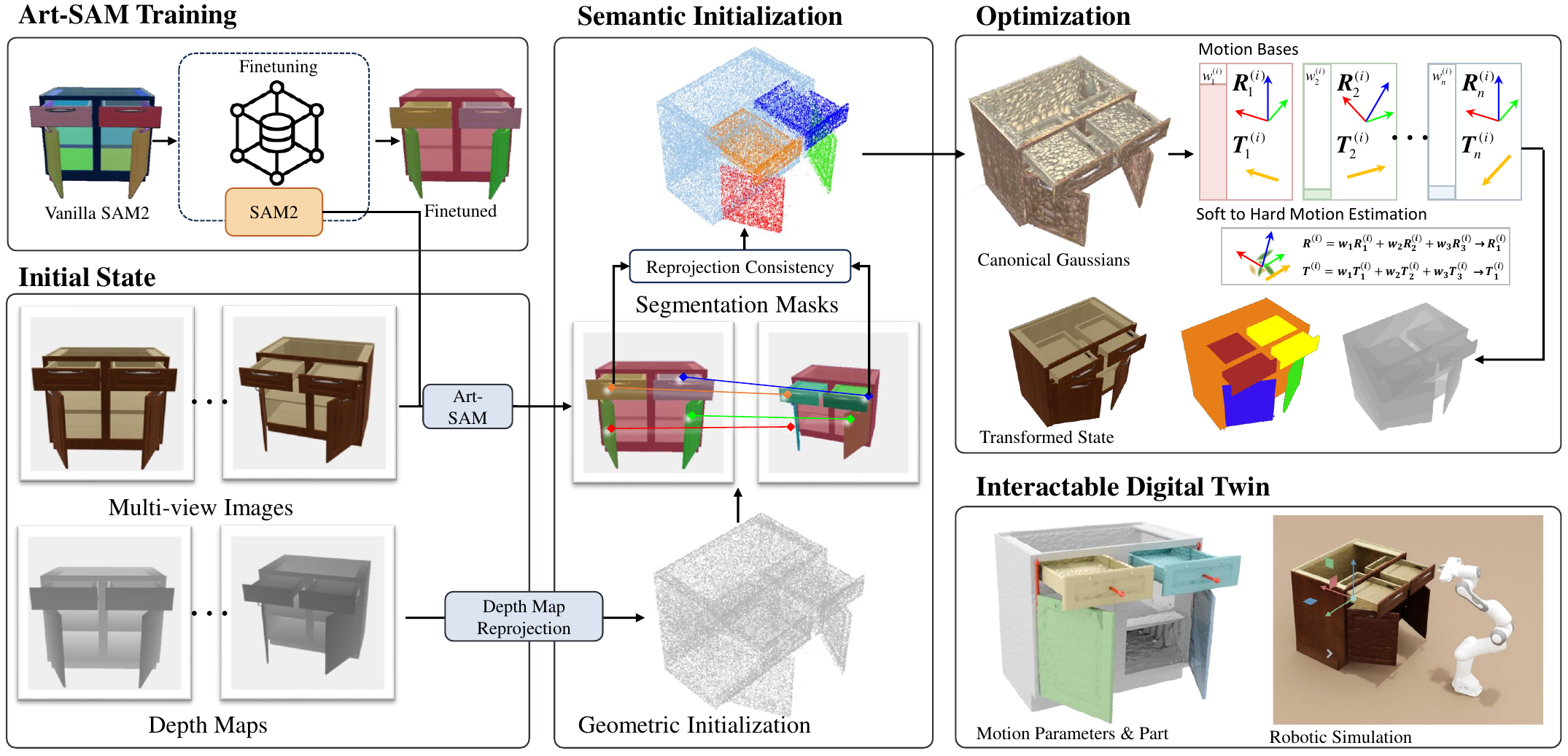}
\vspace{-3mm}
\caption{The overview of GaussianArt. We first design a pipeline to generate multi-view-consistent part segmentation masks, which are used to initialize Gaussians in the canonical state. During training, we introduce a unified framework that jointly learns part segmentation and motion using Gaussians. This process employs a soft-to-hard motion optimization strategy, supervised by RGB-D data and part segmentation masks, along with additional refinement techniques (see \cref{sec:4.3}). Finally, the mesh and motion parameters produced by GaussianArt can be effectively applied to robotic simulation.
}
\label{Fig:overview} 
\vspace{-3mm}
\end{figure*}

\section{Related Work}
\subsection{3DGS and Dynamic Variants}
3D Gaussian Splatting (3DGS) \cite{kerbl20233dgs,song2024sa,yang2024spectrally,zhang2024drone,wang2026unifying,ye2025gs,xu2025cruise} represents 3D scenes using ellipsoids with Gaussian distributions as geometric primitives. This representation enables novel view rendering through differentiable rasterization, making it highly efficient for both training and rendering. The superior performance of 3DGS has inspired a series of studies that apply it to more complex dynamic scenes, exploring various motion modeling approaches \cite{Yang_2024_Deformable, yang2023gs4d, Huang_2024_SCGS, Lin_2024_Gaussian_Flow, wan2024templatefree, xu2024longvolcap}. Several studies, such as GART \cite{lei2024gart}, utilize articulated motion primitives for modeling rigid human body movement via linear blend skinning (LBS), while Shape-of-Motion \cite{som2024} applies a similar model for reconstructing dynamic scenes from monocular videos. Other approaches, like SC-GS \cite{Huang_2024_SCGS}, use sparsely distributed control points with $\mathbf{SE}(3)$ motion priors, allowing user editing of motion, and Gaussian-Flow \cite{Lin_2024_Gaussian_Flow} employs a dual-domain deformation model considering frequency-domain transformations. However, most of these methods lack motion constraints and struggle to accurately estimate part-level articulated motion parameters.


\subsection{3D Articulated Object Modeling}
With the advancements of shape and dynamic modeling \cite{li2025megasam,jones2020shapeassembly,lu2023jigsaw,huang2011joint,tulsiani2016learning,tulsiani2017learning,zou20173d},
many works begin leveraging 3D point clouds \cite{Yi_2018_Deep_Part, wang2019shape2motion, Li_2020_Category_Level, yan2020rpm, Weng_2021_Captra, geng2023sage, geng2023gapartnet,kreber2025guiding,zhong20233d,zhong2022snake}, or images and videos (sometimes with depth) \cite{Jiang_2022_Opd, Qian_2022_understanding_3d_internet, kawana_2023_detection_based_part_level, geng2023sage, Sun_2024_OPDMulti, chen2024urdformer, mandi2024real2code, liu2024singapo,Li_2025_LocateNRotate, zhang2025iaao, xia2025drawer,wu2025dipo,lu2025dreamart,cao2025physx,wu2025predict,mo2021where2act,gao2025partrm} to model articulated objects. However, the most effective way is to reconstruct high-fidelity digital twins through multi-state observation \cite{Mu_2021_ASDF, Jiang_2022_ditto,Tseng_2021_CLA-NeRF,liu2023paris,weng2024digitaltwin, swaminathan24leia, deng2024articulate,liu2025artgs,wu2025reartgs,guo2025articulatedgs,lin2025splart,wang2025self,yu2025part,kim2025screwsplat}.

Former works \cite{Mu_2021_ASDF, Jiang_2022_ditto} use different scene representations to reconstruct articulated objects. The emergence of neural radiance fields \cite{mildenhall_2020_nerf,wang2021neus,liu2020neural,liu2024rip,yuan2024slimmerf} and their high rendering fidelity have led to a series of works adopting neural implicit scene representations for articulated object reconstruction.
\cite{Tseng_2021_CLA-NeRF, liu2023paris} are among the first efforts in reconstructing articulated objects based on neural rendering. Although promising results have been achieved for two-part objects, extending these approaches to multi-part scenarios remains challenging and lacks generalizability. 
Recent studies \cite{weng2024digitaltwin, swaminathan24leia, deng2024articulate} have improved the accuracy of motion parameter prediction and demonstrated some capability in multi-part object reconstruction, but they struggle to generalize to objects with more parts and more complex motion combinations. ArtGS \cite{liu2025artgs}, built upon 3DGS, achieves certain improvements on multi-part objects, but its results are highly unstable and show poor generalization in large-scale evaluations.
In this work, we propose a well-designed unified scalable reconstruction pipeline based on articulated Gaussians.
\section{GaussianArt}

In this section, we present GaussianArt, a unified pipeline for modeling articulated objects using 3DGS. \cref{sec:4.1} provides the design and analysis of the articulated Gaussians. We employ a soft-to-hard training paradigm to optimize the Gaussians as rigid parts (\cref{sec:4.3}) and
build robust initialization (\cref{sec:4.2}). An overview of GaussianArt is shown in \cref{Fig:overview}, with key factors discussed in the following sections.

\subsection{Preliminaries}
3DGS \cite{kerbl20233dgs} represents a 3D scene by a set of Gaussian Primitives, each defined
as:
\begin{equation}
    \mathbf{G}(\mathbf{x}) = \exp\left(-\frac{1}{2}(\mathbf{x}- \bm{\mu})^{T}\bm{\Sigma}
    ^{-1}(\mathbf{x}- \bm{\mu})\right),
    \label{eq:gaussian}
\end{equation}
where $x \in \mathbb{R}^{3 \times 1}$ is Gaussian's 3D position in the scene, $\bm
{\mu}\in \mathbb{R}^{3 \times 1}$ is the mean vector, and
$\bm{\Sigma}\in \mathbb{R}^{3 \times 3}$ is the covariance matrix. To ensure positive
semi-definiteness, $\bm{\Sigma}$ is parameterized as
$\bm{\Sigma}= \mathbf{R}\mathbf{S}\mathbf{S}^{T}\mathbf{R}^{T}$, where $\mathbf{R}
\in \mathbb{R}^{3 \times 3}$ is a rotation matrix and $\mathbf{S}\in \mathbb{R}^{3
\times 3}$ is a scaling matrix.

For rendering, 3D Gaussians are depth-sorted, projected, and alpha-blended on the 2D plane to form pixel colors:
\begin{equation}
    \mathbf{C}=\sum_{i=1 }^{n}{T_i \alpha_i\mathbf{c}_i }, \quad T_{i}= \prod_{j=1}^{i-1}\left(1-\alpha_{j}\right),
    \label{eq:alpha-blend}
\end{equation}
where $n$ is the number of contributing 2D Gaussians, $T_{i}$ is the
transmission factor, $\alpha_{j}$ is the opacity and $\mathbf{c}_{i}$ represents the spherical harmonics-based
color of the $i$-th Gaussian. Other attributes, such as depth, normals, and even semantics, can also be rendered in this way.

To reconstruct articulated objects with multiple 1-DoF motion parts, we follow the two-state observation setting in DigitalTwinArt \cite{weng2024digitaltwin}, requiring posed RGB-D sequences of the same scene in two states as input: $\left\{\overline{I}^t_i, \overline{D}^t_i, \overline{E}^t_i, \overline{K}^t_i \right\}_{i=1}^{N_s}$, $t\in\left\{0, 1\right\}$, where $N_s$ denotes the number of images, $\overline{I}^t_i$ the RGB image, $\overline{D}^t_i$ the depth map, $\overline{E}^t_i$ the camera extrinsics, and $\overline{K}^t_i$ the camera intrinsics.. 

\subsection{Articulated Object Representation}
\label{sec:4.1}
While effective for static geometry, vanilla 3DGS cannot represent part-wise rigid motion or ensure cross-state consistency—each state must be modeled independently, which breaks physical plausibility in articulated systems. To address this, we extend 3DGS with an articulated formulation.

Since 3DGS is an explicit scene representation, rigid motions can be modeled by directly transforming Gaussian primitives. Thus, we reconstruct articulated objects as a motion field over canonical Gaussians. For each primitive, rigid motion applies to its mean and covariance:
\begin{equation}
\widetilde{\mu}^{(i)} = \mathbf{R}^{(i)}\mu^{(i)}+\mathbf{T}^{(i)}, 
\widetilde{\mathbf{\Sigma}}^{(i)} = \mathbf{R}^{(i)}\mathbf{\Sigma}^{(i)}{\mathbf{R}^{(i)}}^T,
\label{eq:transform}
\end{equation}
where $\widetilde{\mu}^{(i)} \in \mathbb{R}^3, \widetilde{\mathbf{\Sigma}}^{(i)}\in\mathbb{R}^{3\times3}$ denotes the means and covariance after transformation, and $\mathbf{R}^{(i)}, \mathbf{T}^{(i)}$ denotes the rigid motion parameters of each Gaussian primitive.
In articulated scenes, the number of movable parts is far fewer than primitives. We therefore define global motion bases $\{\mathbf{R}_i, \mathbf{T}_i\}_{i=1}^N$ ($N$ stands for the number of parts), and express per-Gaussian motion as a weighted combination:
\begin{equation}
    \mathbf{R}^{(i)} = \sum_{j=1}^N w_j^{(i)}\mathbf{R}_j, \mathbf{T}^{(i)} = \sum_{j=1}^N w_j^{(i)}\mathbf{T}_j,
    \label{eq:soft-motion}
\end{equation}
where blending weights $\mathbf{w}^{(i)} = {w_1^{(i)}, w_2^{(i)}, \dots, w_N^{(i)}} \in \mathbb{R}^N$ denote the probability of each primitive belonging to a motion base.

This modeling alone is insufficient for accurately capturing the characteristics of articulated objects due to the lack of constraints. To precisely model articulated objects, the following properties should be satisfied during training:

\begin{itemize}
    \item One-hot weights: each Gaussian strongly belongs to one motion base. 
    \item Spatial sparsity: motion weights vary only near part boundaries.
    \item Rigid estimation: primitives within a part are treated as a rigid body for efficient joint optimization.
\end{itemize}

\subsection{Soft-to-hard Training Paradigm}
\label{sec:4.3}
This formulation encodes both geometry and motion in a differentiable manner and ensures physical consistency across object states. However, optimizing such a soft, over-parameterized motion field is challenging, especially when parts are ambiguous or occluded. To address this, GaussianArt adopts a soft-to-hard training paradigm that learns motion as rigid parts.



During the initial 6,000 iterations, we warm up the Gaussians at the canonical state. After the warm-up, since part geometry, assignments, and motion parameters are initially suboptimal, we employ a soft learning strategy to refine Gaussian weights $\mathbf{w}^{(i)}$ and refine part geometries, which ensures rigid part segmentation while laying the foundation for hard training. Specifically, we estimate Gaussian motion as a soft mode \cref{eq:soft-motion} for smooth learning and use two-state part segmentation masks (\cref{sec:4.2}) to guide $\mathbf{w}^{(i)}$ via rasterization, enforcing robust boundary constraints.

\begin{figure}[t]
    \centering
    \begin{subfigure}[b]{0.45\textwidth}
        \includegraphics[width=\textwidth]{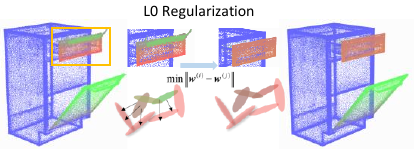}
        \caption{}
        \label{fig:sub1}
    \end{subfigure}
    \hfill
    \begin{subfigure}[b]{0.40\textwidth}
        \includegraphics[width=\textwidth]{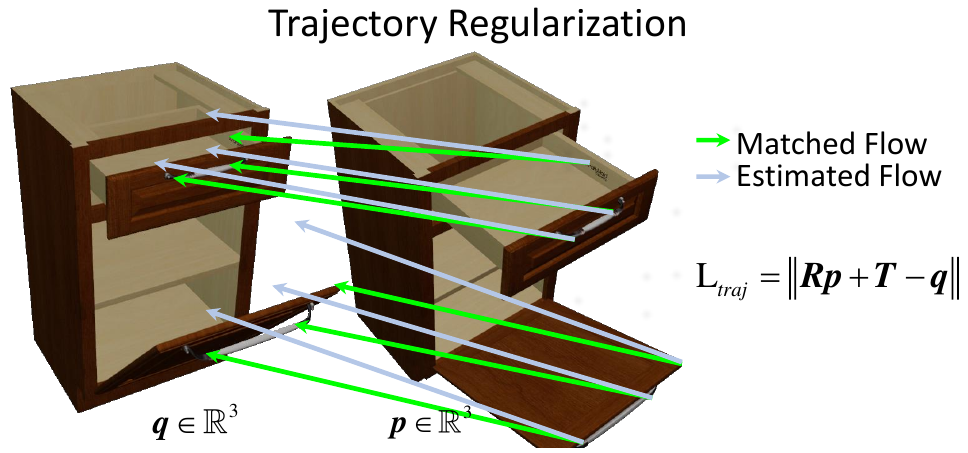}
        \caption{}
        \label{fig:sub2}
    \end{subfigure}
    \vspace{-2mm}
    \caption{Regularization during training. (a) $L_0$ regularization: By using $L_0$ regularization, erroneously assigned Gaussians can be progressively corrected. (b) Trajectory regularization: The point transformed by $\mathbf{p}$ using the estimated flow is constrained to approximate the matched point, facilitating the efficient optimization of motion parameters.}
    \vspace{-5mm}
    \label{fig:regularization}
\end{figure}

However, segmentation masks alone are insufficient for regularizing the weights of all Gaussians. To enhance regularization, we incorporate $L_0$ gradient sparsity during optimization:
\begin{equation}
 \mathcal{L}_{\text{sparsity}} = \sum_{i=1}^N \sum_{j\in \texttt{KNN}(i)} \left\Vert \mathbf{w}^{(i)} - \mathbf{w}^{(j)}\right\Vert.
 \label{eq:sparsity}
\end{equation}
As depicted in \cref{fig:sub1}, $L_0$ regularization corrects misallocated Gaussians, preventing mix-up. It enforces spatial consistency, ensuring nearby Gaussians produce similar predictions and approximate rigid parts, facilitating subsequent hard training.

In other training settings, the Gaussian position learning rate decays exponentially to zero before hard training, ensuring geometric stability and focusing on motion learning. To prevent interference in prismatic parts, we classify a part as prismatic if its rotation remains below a threshold $\epsilon$ after 1,000 soft training steps. We then fix its rotation quaternion to the identity quaternion and detach it from further updates.

During hard training, we disable $L_0$ regularization and assign each Gaussian's motion parameters to those of the part with the highest weight:
\begin{equation}
\mathbf{R}^{(i)} = \mathbf{R}_{j*},\mathbf{T}^{(i)}=\mathbf{T}_{j*},\text{where } j^* = \arg\max_j w_j^{(i)}.
\label{eq:assign}
\end{equation}
This enables simple and direct motion optimization. 


\begin{figure*}[t]
    \centering
    \includegraphics[width=0.95\linewidth]{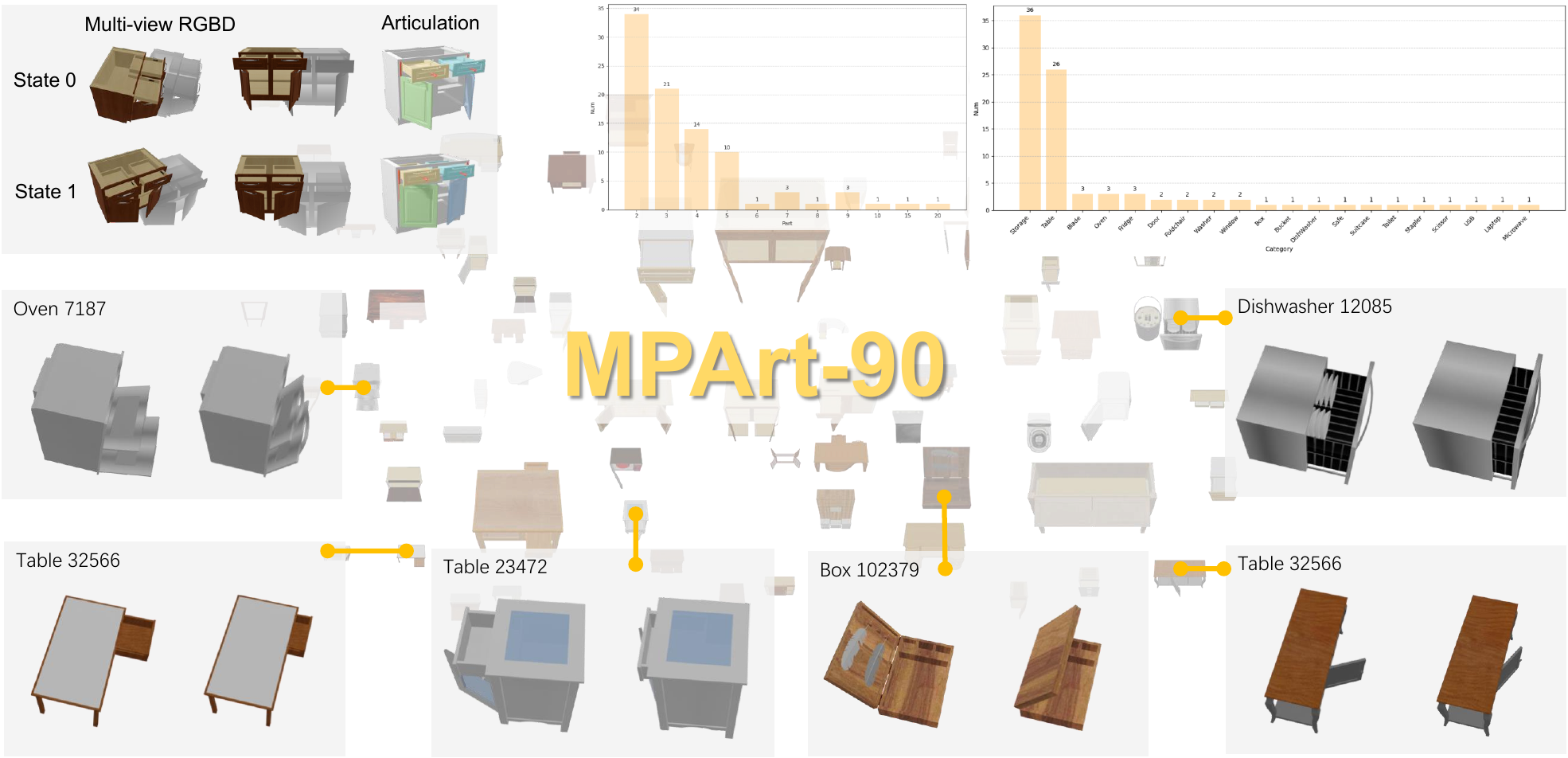}
    \vspace{-2mm}
    \caption{MPArt-90 benchmark. Unlike prior datasets that contain fewer than 20 objects and quickly saturate, MPArt-90 scales articulated object reconstruction to 90 objects across 20 diverse categories. Each object provides multi-view RGBD observations together with ground-truth motion parameters, covering configurations with up to 20 parts. This scale reveals failure cases of prior methods, which often collapse beyond 2–3 parts due to brittle initialization. By offering a large and physically grounded benchmark, MPArt-90 enables systematic evaluation of scalability and generalization in articulated modeling.}
    \vspace{-3mm}
    \label{fig:MPArt-90}
\end{figure*}

Sometimes, part segmentation alone cannot fully constrain motion parameter learning, making extreme movements and occlusions difficult to model. To refine motion learning, we apply a feature-matching-based trajectory regularization term during hard training. For image $I_{v}^0$ from view $v$ at state $0$, we select the top $k$ closest view from state $1$ and apply an image feature matching model \cite{sun2021loftr} to establish 2D pixel correspondences $\{(\mathbf{p}_i, \mathbf{q}_i)\}_{i=1}^{N_m}$, where $N_m$ denotes the number of matches and $\mathbf{p}_i, \mathbf{q}_i$ are pixel coordinates. We lift them to 3D space as $\mathcal{M} = \{(\tilde{\mathbf{p}}_i, \tilde{\mathbf{q}}_i)\}_{i=1}^{N_m}$ using depth and camera parameters. Then, we apply a 3D locality filter on 3D maches to mitigate false correspondences, yielding a refined set of matching pairs $F(\mathcal{M})$. The transformation $(\mathbf{R}_j, \mathbf{T}_j)$ at the matching positions can be derived from \cref{eq:assign}. The trajectory regularization term is the average difference between the position obtained through transformation and that derived from the filtered correspondences (\cref{fig:sub2}):
\begin{equation}
 \mathcal{L}_{traj} = \sum_{j\in F(\mathcal{M})}\left\Vert(\mathbf{R}_j\mathbf{p}_j+\mathbf{T}_j) - \mathbf{q}_j\right\Vert.
 \label{eq:traj}
 \end{equation}

\subsection{Optimization}
\label{sec:4.4}
We supervise training via a multi-term loss that balances appearance fidelity, part consistency, and physically plausible motion. $\mathcal{L}_{\text{RGB-D}}$ is the rendering loss like:
\begin{equation}
\mathcal{L}_{\text{RGB-D}}=(1-\lambda_{\text{SSIM}})\mathcal{L}_1+\lambda_{\text{SSIM}}\mathcal{L}_{
\text{D-SSIM}}+\lambda_{\text{D}}\mathcal{L}_{\text{D}}
\label{eq:rgbd}
\end{equation}
where $\mathcal{L}_1=\left\Vert I-\overline{I}\right\Vert_1$, $\mathcal{L}_{
\text{D-SSIM}}$ is the D-SSIM loss \cite{kerbl20233dgs}, and $\mathcal{L}_{\text{D}}=\left\Vert D-\overline{D}\right\Vert_1$. We also use the segmentation loss:
\begin{equation}
\mathcal{L}_{\text{SEM}} = \text{H}(P, \overline{P})
\label{eq:sem}
\end{equation}
where $P$ is the rendering of weights $\mathbf{w}$, $\overline{P}$ is the segmentation mask of parts, and H is the cross-entropy loss. All in all, our supervision could be summarized as:
\begin{equation}
\mathcal{L}_{\text{Soft}}=\mathcal{L}_{\text{RGB-D}}+\lambda_{\text{SEM}}\mathcal{L}_{\text{SEM}}+\lambda_{\text{sparsity}}\mathcal{L}_{\text{sparsity}},
\label{eq:soft}
\end{equation}
\begin{equation}
\mathcal{L}_{\text{Hard}}=\mathcal{L}_{\text{RGB-D}}+\lambda_{\text{SEM}}\mathcal{L}_{\text{SEM}}+\lambda_{\text{traj}}\mathcal{L}_{\text{traj}},
\label{eq:hard}
\end{equation}
where \cref{eq:soft} is for soft training, and \cref{eq:hard} is for the hard. See Supplementary Material \cref{sec:11} for details.

\subsection{Initialization}
\label{sec:4.2}

\noindent \textbf{Part segmentation.} 
To initialize and regularize Gaussian part weights $\mathbf{w}^{(i)}$, we leverage SAM2 \cite{ravi2024sam2}, a foundation segmentation model pretrained on large-scale data. Since its zero-shot results often show inconsistent granularity on articulated objects, we fine-tune it on multi-view images and masks rendered from Partnet-Mobility (PM) \cite{Xiang_2020_SAPIEN}, yielding a specialized model, Art-SAM. We then apply cross-view propagation to ensure multi-view consistency (see Supplementary Material \cref{sec:9,sec:10}). This method is robust and generalizable, as Art-SAM requires only light post-training to adapt to novel object categories or configurations.

\noindent \textbf{Canonical Gaussians Initialization.}
First, we select the joint state with higher visibility as the canonical state. After obtaining view-consistent segmentation masks, we randomly sample points from RGB-D images and reproject both the color and the part label into 3D space using the depth map. This process yields point clouds for initializing the canonical Gaussians. The part label $S^{(i)}$ is then employed as an affinity feature $ \mathbf{w}^{'(i)} = (w_1^{'(i)}, w_2^{'(i)},\dots, w_N^{'(i)}) \in \mathbb{R}^N$ attached to the Gaussian and serves as the initialization of the weights term $\mathbf{w}^{(i)}$ in motion estimation as follows:
\begin{equation}
    w_j^{'(i)} = \begin{cases} 
    1 & \text{if } j = S^{(i)}, \\
    0 & \text{if } j \neq S^{(i)},
    \end{cases}
    \label{eq:init1}
\end{equation}
\begin{equation}
    \mathbf{w}^{(i)}=\text{Softmax}(\mathbf{w}^{'(i)}).
    \label{eq:init2}
\end{equation}
This method for part initialization is highly robust while also providing ample room for modification.

\section{MPArt-90 Benchmark}

\subsection{Benchmark Data Generation}
We have constructed a novel benchmark, MPArt-90, containing 90 objects from 20 categories, extending the existing dataset for articulated object reconstruction to a larger scale, covering more object types and motion patterns, as shown in \cref{fig:MPArt-90}. Compared to earlier benchmarks limited to fewer than 20 objects and only 2–3 part articulations, MPArt-90 emphasizes diversity and realism, exposing failure modes in brittle pipelines and enabling systematic evaluation of generalization and robustness.

Articulated objects are mainly constructed from the PM dataset \cite{Xiang_2020_SAPIEN}, from which we select 87 objects based on the diversity of categories, part number, and appearances. We use Blender with a procedural rendering pipeline \cite{blender, Denninger2023blenderproc2} to render multi-view images of the 3D models in the base assets. To increase the diversity of object states while providing adequate observation of the interior parts of the objects, we set the 1-DoF part-level motion parameter to two random states: the starting state lies between $[0.65, 0.75]$ and the ending state is within $[0.35, 0.45]$ (here $0$ denotes the "fully-closed" state and $1$ denotes the "fully-open" state). For the object at each motion state, we place the camera in a spherical region around the object and randomly sample 100 views for training and 20 views for testing, all at a resolution of $800\times800$. 

Due to the limited availability of high-quality real-world articulated object data and the significant ground-truth errors caused by the inability to annotate the internal structures of real objects, we selected only three well-annotated real objects from the Multiscan \cite{mao2022multiscan} dataset.
See Supplementary Material \cref{sec:18} for more results.

\sisetup{detect-all=true}
\begin{table*}[t]
\centering
\def\mywidth{0.95\textwidth} 
\def\myspc{}{\hspace{8ex}}
\resizebox{\mywidth}{!}{
\sisetup{table-auto-round,table-format=.2, table-column-width=1.35cm}
\begin{tabular}{cl|SSSSS}

\hline
                               &                        & \multicolumn{1}{c}{2 Parts (34)}                      & \multicolumn{1}{c}{3 Parts (21)}                         & \multicolumn{1}{c}{4-5 Parts (24)}   & \multicolumn{1}{c}{6-20 Parts (11)}                  & \multicolumn{1}{c}{All  (90)\bigstrut[b]}\\
\hline

\hline
\multirow{2}[2]{*}{\shortstack{Axis Ang}}       

& ArtGS \cite{liu2025artgs} & \bfseries 3.53 &11.61 & 15.49 & 35.66 & 24.34\bigstrut[b]\\

& Ours &
 4.90 &\bfseries6.33 &\bfseries 12.43 &\bfseries 12.05 &\bfseries 12.17 \bigstrut[b]\\

\hline
\multirow{2}[2]{*}{\shortstack{Axis Pos}}       

& ArtGS \cite{liu2025artgs} & 1.08 & 1.62 &\bfseries 1.09 & 4.62 & 1.45\bigstrut[b]\\

& Ours 
&\bfseries 0.27 &\bfseries 0.38 & 1.30 &\bfseries3.06 &\bfseries 1.06\bigstrut[b]\\

\hline
\multirow{2}[2]{*}{\shortstack{Part Motion}}       

& ArtGS \cite{liu2025artgs} & 7.98&\bfseries6.08 &\bfseries 4.45& 13.74& 10.16 \bigstrut[b]\\

& Ours 
&\bfseries 7.57 & 11.82& 9.03&\bfseries 7.14 &\bfseries 9.07\bigstrut[b]\\

\hline
\multirow{2}[2]{*}{CD-s}       

& ArtGS \cite{liu2025artgs} & 5.39 & 13.37  &18.02 &13.14 &11.57   \bigstrut[b]\\

& Ours 
&\bfseries 2.75&\bfseries2.47&\bfseries 2.31 &\bfseries3.7 &\bfseries 2.68\bigstrut[b]\\

\hline

\multirow{2}[2]{*}{CD-m}       

& ArtGS \cite{liu2025artgs} &47.80 &194.15 & 340.53  &459.81 &380.29\bigstrut[b]\\

& Ours &\bfseries 4.61&\bfseries6.17 &\bfseries 5.42 &\bfseries5.43 &\bfseries 5.46\bigstrut[b]\\
\hline

\end{tabular}
}
\vspace{-2mm}
\caption{Quantitative results on MPArt-90 benchmark. Metrics are shown as the mean $\pm$ std over 3 trials with different random seeds following \cite{weng2024digitaltwin}.
}
\label{tab:MPArt-90}
\end{table*}
\begin{figure*}[t] 
\centering 
\includegraphics[width=0.95\textwidth]{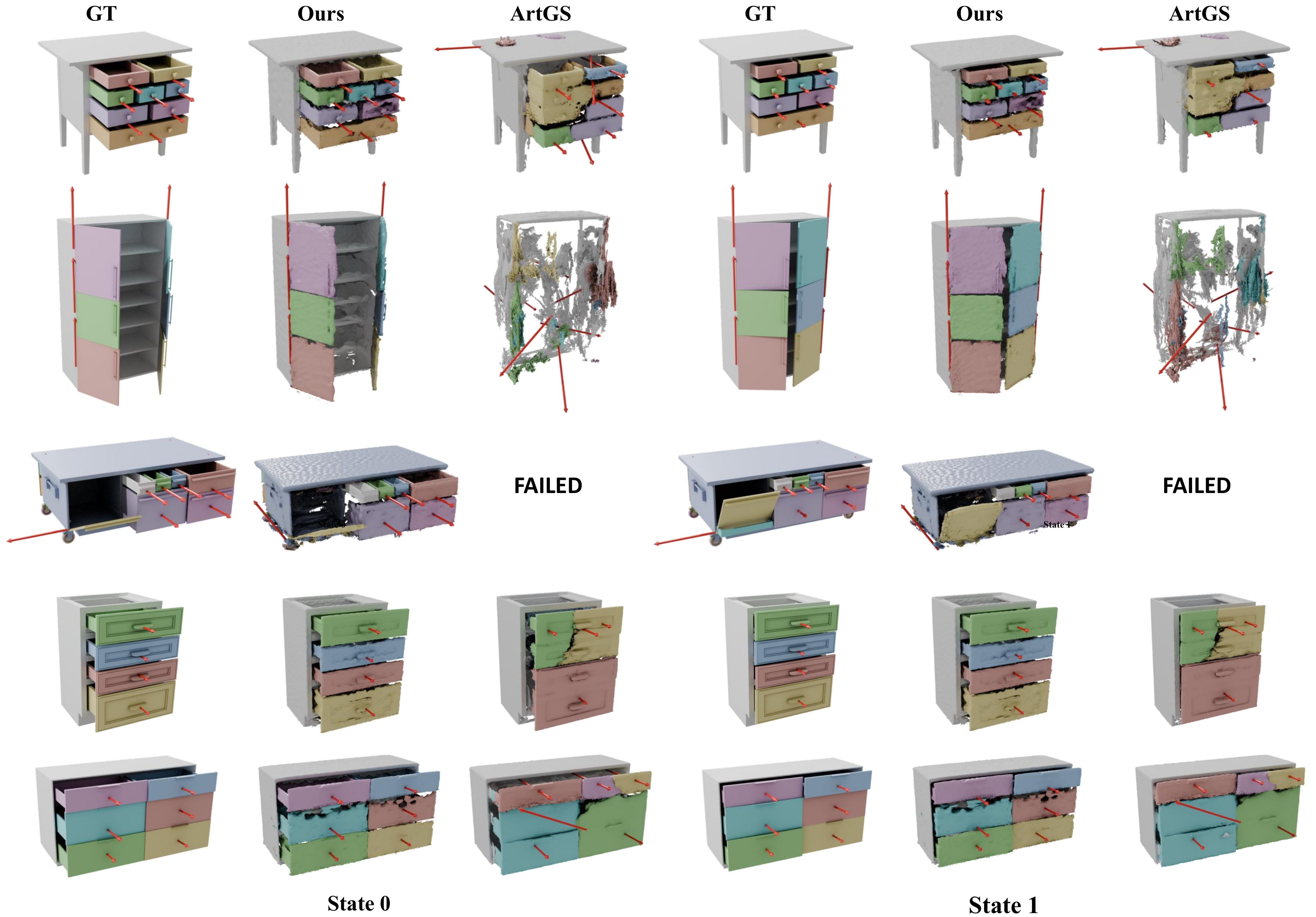}
\vspace{-2mm}
\caption{Qualitative results on multi-part objects of MPArt-90.}
\label{Fig:reconvis} 
\vspace{-3mm}
\end{figure*}

\section{Experimental Results}
\label{sec:result}
\subsection{Implementations}
We first evaluate several methods on several traditional datasets, and finally select our method and ArtGS \cite{liu2025artgs} for scale-up evaluation on our MPArt-90 benchmark. 

For metrics, we calculate Axis Pos Error, Axis Angle Error, and Part Motion Error for motion parameters estimation, and  CD for static and dynamic part separately for geometric reconstruction. We report the mean for each metric over the 3 trials at the high-visibility state. See Supplementary Material \cref{sec:15,sec:16} for more details.

\subsection{Experiments on MPArt-90}


As depicted in \cref{tab:MPArt-90}, when the object contains fewer parts, ArtGS performs comparably to, or even slightly better than, GaussianArt. However, as the number of parts increases, the stability of ArtGS degrades and the performance gap between the two methods widens. This is particularly evident in the geometry of dynamic parts, where ArtGS’s cluster-based initialization lacks strong generalizability. In many cases, it fails to achieve effective part segmentation, resulting in significant errors in motion parameter estimation. In contrast, GaussianArt uses a fine-tuned vision foundation model and a multi-view consistency pipeline to produce high-quality part segmentation, providing stable initialization for motion parameter learning and improving generalization to diverse objects.

Moreover, as illustrated by the qualitative results in \cref{Fig:reconvis}, when handling objects with multiple parts exhibiting similar motion patterns, ArtGS often produces ambiguous segmentation results, which in turn lead to inaccurate motion parameter estimation. Such errors frequently cause parts to split during motion, adversely affecting the overall reconstructed geometry. In more complex scenarios, such as objects comprising more than 20 parts, ArtGS may even fail to converge during training. In contrast, GaussianArt exhibits strong generalization capabilities: even when confronted with highly articulated objects, it consistently delivers accurate part segmentation and reliable motion parameter predictions. These strengths enable the reconstruction of high-fidelity digital twins, laying a solid foundation for wider deployment in downstream applications. See Supplementary Material \cref{sec:17} for more results.

\subsection{Ablation Studies}
\begin{table}[h]
\centering
\def\mywidth{0.50\textwidth} 
\resizebox{\mywidth}{!}{
\sisetup{table-auto-round,table-format=.2,table-column-width=1.5cm}
\begin{tabular}{l|ccccc}
\hline
                               & \multicolumn{1}{c}{Axis Ang}                       & \multicolumn{1}{c}{Axis Pos}                       & \multicolumn{1}{c}{Part Motion}                    & \multicolumn{1}{c}{CD-s}                           & \multicolumn{1}{c}{CD-m}
                               \bigstrut\\
\hline
Proposed &\bfseries 0.03 &\bfseries 0.01 &\bfseries 0.04 &\bfseries 0.67
&\bfseries 0.14 \bigstrut[t]\\
w/o Part-seg & 28.60 & 4.67 & 18.50 & 0.88 & 186.67 \\
w/o $L_0$ & 0.10 & 0.01 & 0.08 & 0.70 & 0.17 \\
w/o Traj & 0.31 & 0.03 & 0.35 & 0.85 & 0.25 \\
w/o Part-init & 0.25 & 0.02 & 0.15 & 0.71    & 1.34 \\
w\quad MLP Seg & 41.57 & 3.78 & 31.43 & 2.68 & 478.20 \bigstrut[b]\\
\hline
\end{tabular}%
}
\vspace{-2mm}
\caption{Results of ablation studies.}
\label{tab:ablation}
\vspace{-5mm}
\end{table}

\noindent\textbf{Implementation Details.} To evaluate the effectiveness of different components, we design several ablation studies on 5 multi-part objects. All metrics are averaged across 5 trials. 

\noindent\textbf{Results.} The results are as follows:
\begin{itemize}
    \item\textit{Part assignment.} We evaluate our model under 3 different scenarios: (1) without part segmentation masks, (2) without part initialization, and (3) replacing the part assignment module with an MLP, as used in GART \cite{lei2024gart}. As shown in Table~\ref{tab:ablation}, without segmentation masks, Gaussians struggle to capture the complex motion of multiple parts. Additionally, relying solely on part segmentation supervision—without explicit part initialization—can negatively impact the geometric reconstruction of movable parts. When using MLPs for part assignment, the model fails entirely to learn part segmentation and motion for multi-part objects.

    \item\textit{Trajectory regularization.}  As demonstrated in Table~\ref{tab:ablation}, trajectory regularization enhances part motion learning, improving articulated reconstruction.

    \item\textit{$L_0$ regularization.} As observed in Table~\ref{tab:ablation}, $L_0$ regularization refines part assignment, leading to more accurate part articulation modeling.
\end{itemize}
\begin{figure*}[t] 
\centering 
\includegraphics[width=0.95\textwidth]{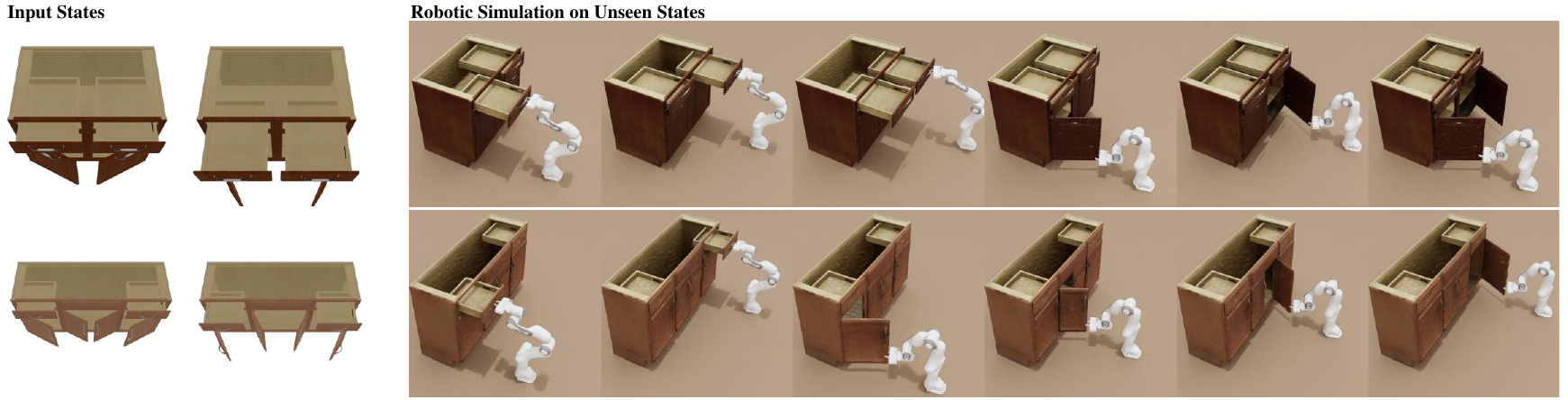}
\vspace{-2mm}
\caption{Digital twins reconstructed by GaussianArt in NVIDIA Omniverse IssacSim.}
\label{Fig:isaac} 
\vspace{-3mm}
\end{figure*}

\section{Application}
\subsection{Robotic Manipulation}
\cref{Fig:isaac} presents GaussianArt's reconstruction of multi-part articulated objects in NVIDIA Omniverse IssacSim. Leveraging learned motion parameters and precise part-level geometry, we effectively decomposed the hybrid motion in visual observations, enabling a robot arm to interact with any moving part at unseen states in input images. These realistic digital twins facilitate robotic manipulation of articulated objects.

\subsection{HSI}
\begin{figure}[t] 
\centering 
\includegraphics[width=0.45\textwidth]{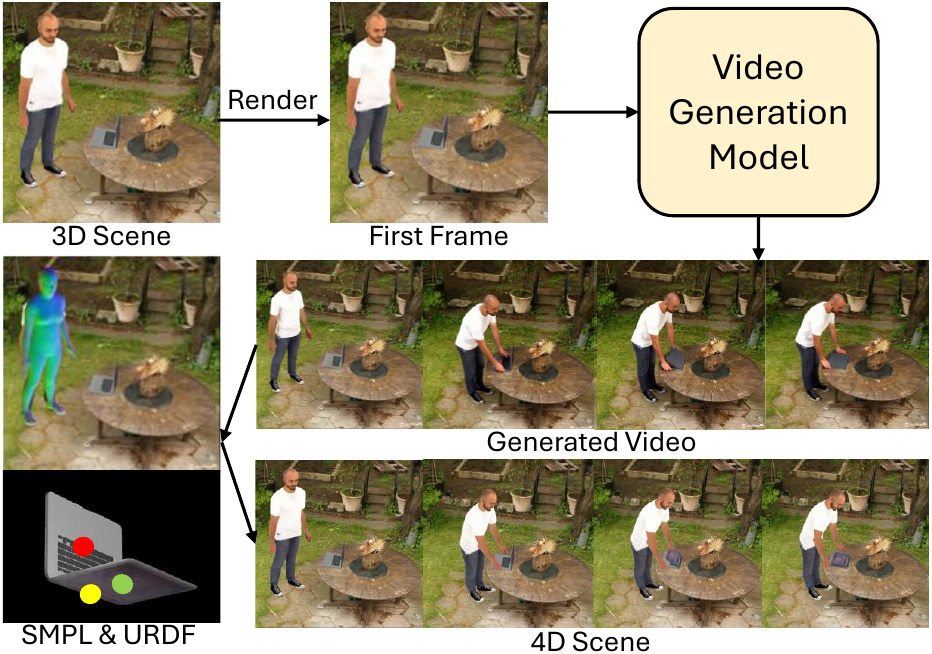}
\vspace{-3mm}
\caption{HSI pipeline with the digital twin generated by GaussianArt.}
\label{Fig:HSI} 
\vspace{-5mm}
\end{figure}

Furthermore, our Articulated Gaussians can be leveraged to generate 4D assets, enabling the modeling of HSI in dynamic environments. As depicted in \cref{Fig:HSI}, given Gaussian-based representations of humans, objects, and scenes as input, inspired by ZeroHSI \cite{li2024zerohsi}, we could generate high-fidelity 4D scenes. We first render a static frame, which is then used as input to a text-guided video diffusion model to synthesize a corresponding video. This generated video serves as a supervisory signal to optimize the kinematic parameters of humans and objects, effectively lifting the original 3D assets into a coherent 4D representation.

Specifically, we employ SMPL \cite{loper2023smpl} for human kinematic modeling and use our reconstructed digital twins, with articulation parameters, to represent objects with kinematics. Through a distillation process, the motion dynamics captured in the video are transferred to the 4D scene.

We believe this approach holds strong potential for a wide range of applications and will offer significant value across multiple domains in the future.

\section{Limitations}
Despite its strengths, GaussianArt has several limitations. First, the lack of direct constraints on intermediate motion states can lead to incorrect motion parameter learning, particularly in extreme transitions (e.g., from a fully open to a fully closed door). Future work will explore constraint-based strategies within the unified GS framework to improve motion learning.
Second, the initialization of canonical Gaussians may be suboptimal due to out-of-distribution issues in part segmentation or misalignments in multi-view reconstruction. These imperfections can negatively impact the learning of motion parameters. Addressing this challenge will require developing more robust multi-view segmentation methods. See Supplementary Material \cref{sec:14} for more details.

\section{Conclusion}
\label{sec:conclusion}
In this work, we introduce GaussianArt, a unified modeling pipeline for articulated objects. Our approach begins with a robust part segmentation model to initialize canonical Gaussians, followed by a soft-to-hard training paradigm for improved motion optimization. Extensive experiments show that GaussianArt achieves state-of-the-art (SoTA) performance in geometric reconstruction and part motion estimation on our curated largest benchmark for articulated objects reconstruction, MPArt-90. The digital twins created by GaussianArt can be seamlessly integrated into simulators for tasks like articulated object manipulation, and can also be used for modeling human–scene interactions, where we hope to enable more natural and adaptive simulation of real-world environments.
{
    \small
    \bibliographystyle{ieeenat_fullname}
    \bibliography{main}
}

\clearpage
\setcounter{page}{1}
\maketitlesupplementary

\section{Art-SAM Training}
\label{sec:9}
\begin{figure*}[t]
    \centering
    \includegraphics[width=0.95\linewidth]{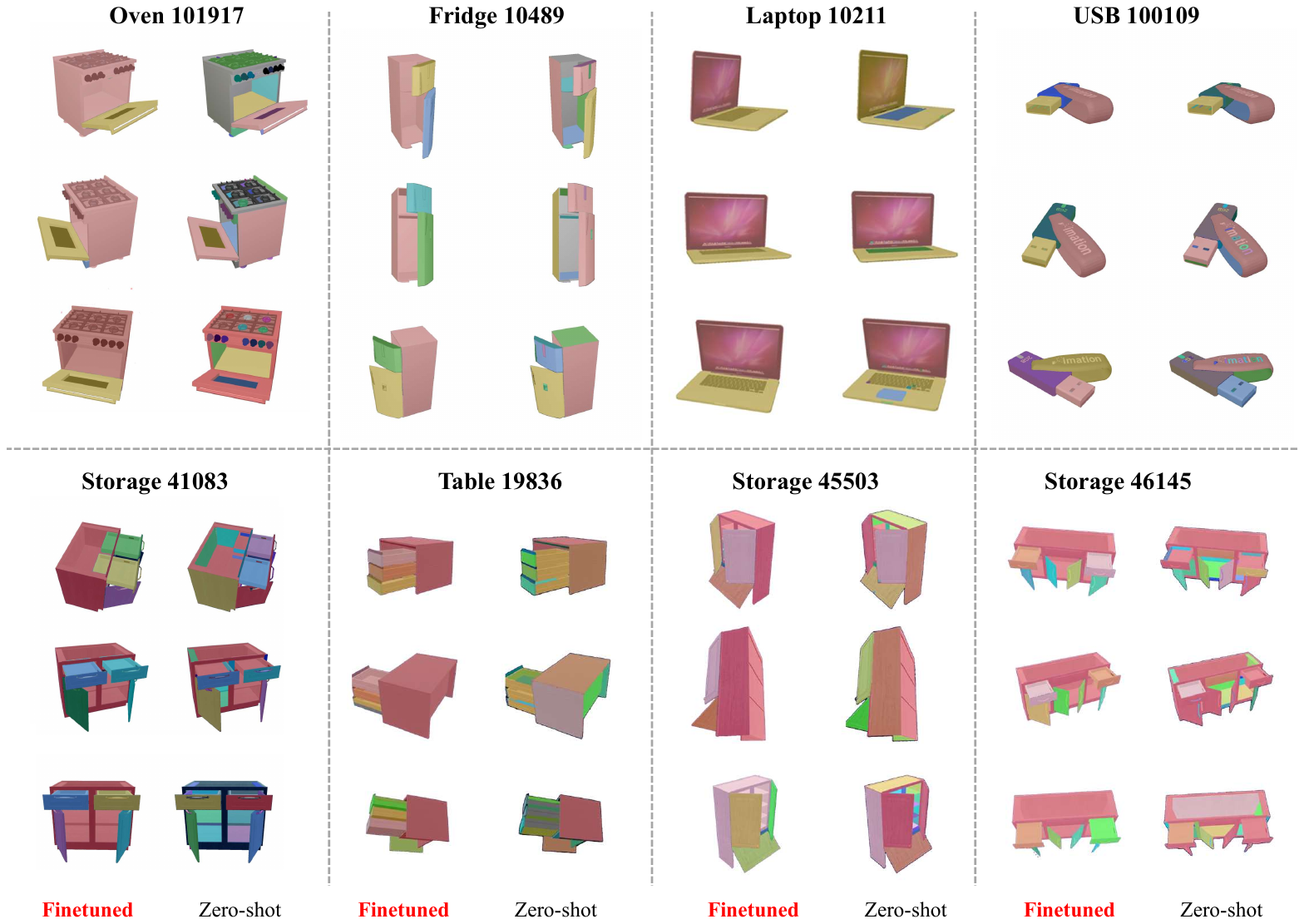}
    \vspace{-2mm}
    \caption{Visual Comparisons of segmentation masks generated by our fine-tuned \textit{\textbf{Art-SAM}} model and Zero-shot SAM-v2.1 model. }
    \label{fig:SAM_supp}
    \vspace{-2mm}
\end{figure*}

\begin{figure}[t]
    \centering
    \includegraphics[width=0.85\linewidth]{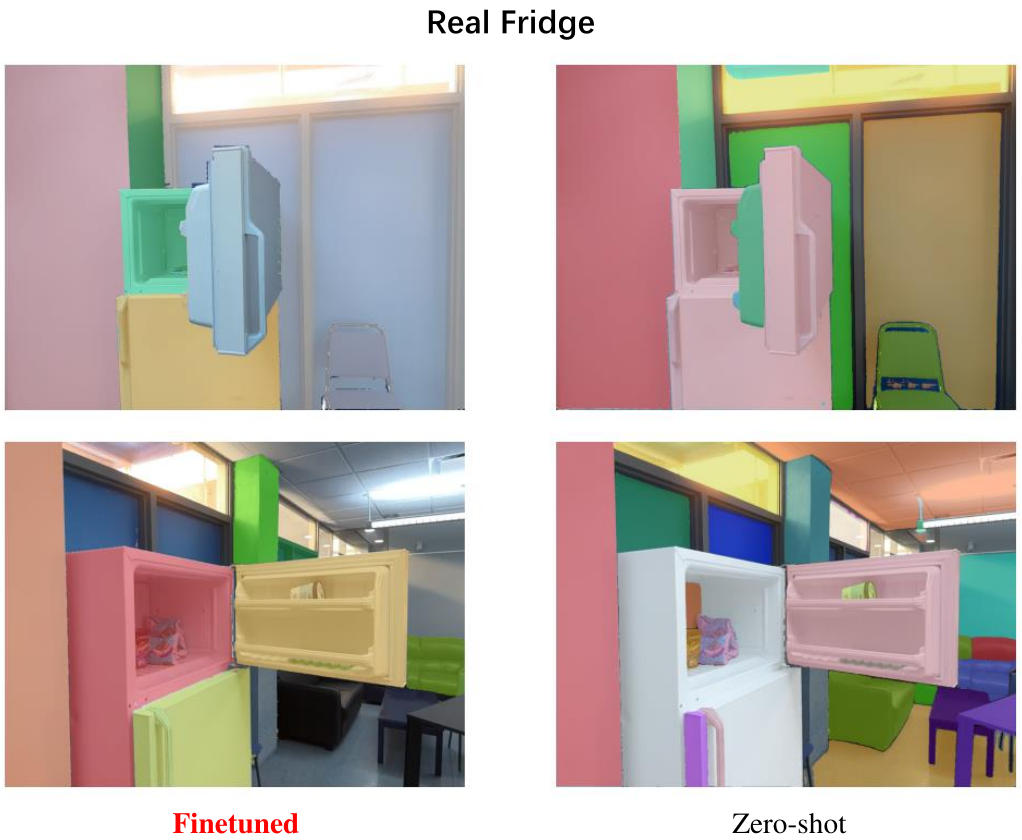}
    \vspace{-2mm}
    \caption{Visual Comparisons of segmentation masks generated by our fine-tuned \textit{\textbf{Art-SAM}} model and Zero-shot SAM-v2.1 model on real objects.}
    \label{fig:realSAM_supp}
    \vspace{-5mm}
\end{figure}

To provide supervision for accurate part-level segmentation, we adopt an image segmentation model to generate segmentation masks for each view and then use multi-view reprojection consistency to propagate monocular masks across views. To this end, an image segmentation model is required. Visual foundation models \cite{kirillov2023segment, ravi2024sam2} have emerged as powerful tools for class-agnostic zero-shot or prompt-based image segmentation. However, when applied to articulated objects, the problems of over-segmentation and part ambiguity are common. Applying the models in a zero-shot manner often results in masks of erroneous granularity, where some masks cover overly fine-grained areas, while others fail to cover the full range of a complete interactable part (see the 'Zero-shot' results of \cref{fig:SAM_supp}). Therefore, we fine-tuned SAM-v2.1 model \cite{ravi2024sam2} to generate segmentation masks for articulated objects at the desired granularity. The fine-tuned model can generate masks of the desired part-level granularity on synthetic objects (\cref{fig:SAM_supp}) and real objects (\cref{fig:realSAM_supp}). This section covers the details of data preparation and the fine-tuning process.

\begin{figure}[htbp]
    \centering
    \includegraphics[width=0.95\linewidth]{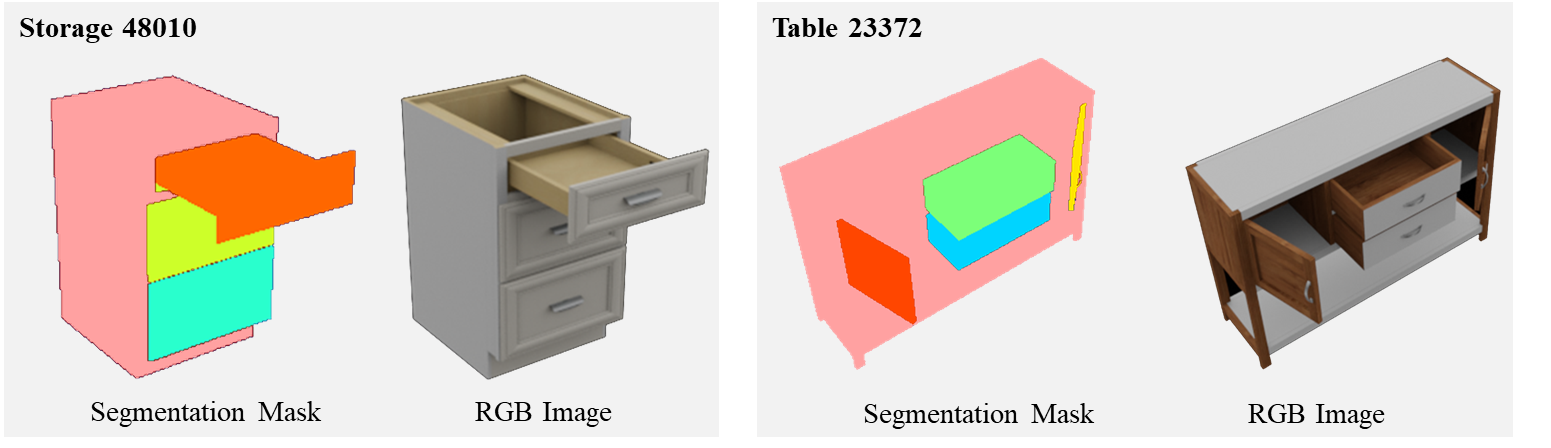}
    \vspace{-3mm}
    \caption{Examples of training data used for fine-tuning}
    \label{fig:training_samples}
    \vspace{-2mm}
\end{figure}

\noindent{\textbf{Base Assets.}} We use the PM dataset \cite{Xiang_2020_SAPIEN} to create a dataset for fine-tuning SAM-v2.1. The dataset contains URDF-format models of articulated objects with motion parameters. We select 207 objects from 20 categories to generate pairs of images and corresponding segmentation masks for fine-tuning. For real-world assets, we select 50 objects from Multiscan \cite{mao2022multiscan}.

\noindent{\textbf{Image Rendering.}} We use Blender with a procedural rendering pipeline \cite{blender, Denninger2023blenderproc2} to render multi-view images of the 3D models in the base assets. To increase the diversity of object states while providing adequate observation of the interior parts of the objects, we set the 1-DoF part-level motion parameter to random states between $[0.2, 0.8]$ (here $0$ denotes the "fully closed" state and $1$ denotes the "fully open" state). For each object, we place the camera on a spherical region around the object and randomly sample 100 to 300 views, generating 130,000 training images at three resolution levels: $512\times512$, $800\times800$, and $1024\times1024$. Examples of training data are shown in \cref{fig:training_samples}.

\noindent{\textbf{Segmentation Map Rendering.}} To provide supervision at the desired granularity, we also adopt the procedural rendering pipeline with Blender \cite{Denninger2023blenderproc2}, which allows the rendering of object masks. A training sample consists of an RGB image paired with its corresponding segmentation masks.

\noindent{\textbf{Training Setup.}} Starting from the pretrained \textit{SAM2.1-Hiera-B+} model, we fine-tune all model parameters for 40 epochs on the dataset we created from PM. The loss function is a weighted linear combination of MSE IoU loss, Dice loss \cite{milletari2016vnet}, and focal loss \cite{Lin_2017_Focal_Loss}. In each epoch, the learning rate starts at $5\times10^{-6}$ and decays to $5\times10^{-7}$ using a cosine annealing scheduler. 

\begin{figure}[t]
    \centering
    \includegraphics[width=0.85\linewidth]{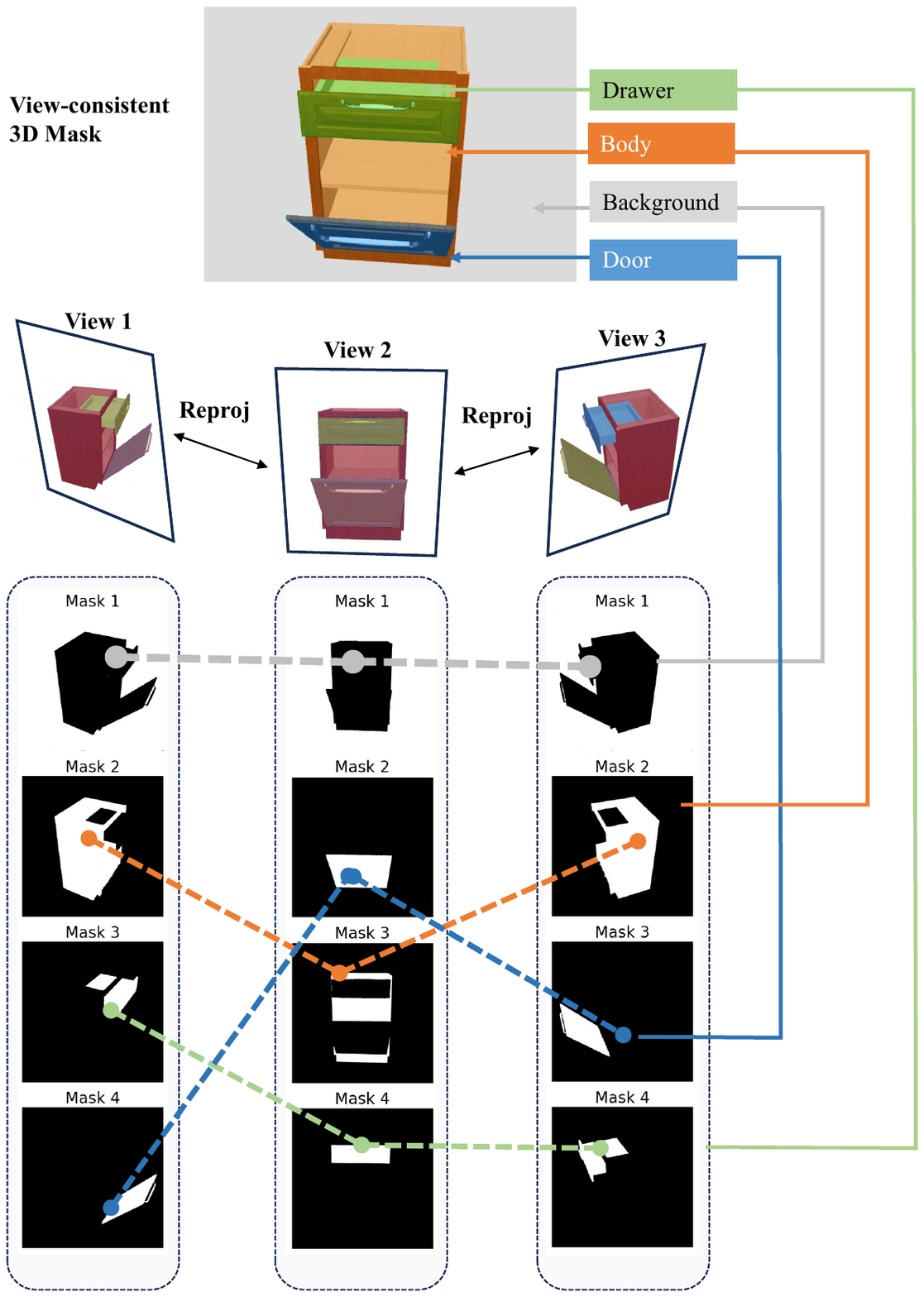}
    \vspace{-3mm}
    \caption{Visual illustration of mask reprojection matching and graph construction.}
    \label{fig:reprojection}
    \vspace{-5mm}
\end{figure}

\section{Multi-view Mask Consistency}
\label{sec:10}
The fine-tuned model can generate masks at the desired granularity, but the masks are inconsistent across views. Some studies \cite{Ying_2024_Omniseg3d, cen2024segment3dgaussians} adopt a contrastive learning strategy to achieve multi-view consistency. A random feature vector is initialized for each Gaussian primitive and optimized using a contrastive loss. At test time, object masks are obtained through density-based clustering \cite{ester1996density} and projected to the target view. This approach is effective for prompt-based segmentation; however, it is not well suited for part-level motion supervision.

In the context of single-object 3D segmentation, monocular masks provide only partial coverage of the target object, making them insufficient for comprehensive observation. To address this limitation, SA3D \cite{cen2023_sa3d} introduced a multi-view mask re-prompting algorithm to mitigate the issue of incomplete observations. Building upon this approach, we extend the mask re-prompting method proposed in SA3D \cite{cen2023_sa3d} to a multi-mask setting to generate multi-view consistent part labels. Given source view $i$ and target view $j$, we have RGB-D images $(\mathbf{I}_i, \mathbf{D}_i), (\mathbf{I}_j, \mathbf{D}_j)$, camera parameters $(\mathbf{K}_i, \mathbf{E}_i), (\mathbf{K}_j, \mathbf{E}_j)$ and segmentation masks generated by Art-SAM: $\{\mathbf{M}_i^{(s)}\}_{s=1}^{N_i}, \{\mathbf{M}_j^{(t)}\}_{t=1}^{N_j}$, where $N_i$ and $N_j$ denote the number of segmentation masks. For each mask obtained from the source view mask, we first obtain a 3D point mask by projecting each of the masked regions with the depth map and camera parameters:

\begin{equation}
\mathbf{p}_i^{(s)} = \mathbf{M}_i^{(s)}\odot\mathbf{D}_i\odot({\mathbf{K}_i}^{-1}{\mathbf{E}_i}),
\end{equation}
\label{eq:14}
\\
where $\mathbf{D}^{(i)} \in \mathbb{R}^{H\times W}$ is the depth map, ${\mathbf{K}_i}$ and ${\mathbf{E}^{(i)}}$ denote intrinsics and extrinsics of the camera, $\odot$ is Hadamard product.

Then, the mask reprojected from view $i$ to $j$ can be calculated by multiplying the 3D point mask $\mathbf{p}_i^{(s)}$ with the camera parameters of the target view:

\begin{equation}
    \mathbf{M}_{i\rightarrow j}^{(s)} = \mathbf{K}_j\mathbf{E}_j^{-1}\mathbf{p}^{(s)}_i
\end{equation}
\label{eq:15}
\\

The confidence score is calculated as the IoU of the re-projected mask with the masks of the target view. The matching relation can be described by a matrix $\mathbf{S}$ whose element at position $s, t$ is:

\begin{equation}
    \mathbf{S}_{s, t} = \textnormal{IoU}\left(\mathbf{M}_{i\rightarrow j}^{(s)}, \mathbf{M}_j^{(t)}\right) 
\end{equation}
\label{eq:16}
\\

Starting with the anchor view, we select the $k$ nearest views and establish correspondences by applying the Hungarian algorithm to pairs of adjacent views. After completing the iterative matching process, a graph representing the correlations between masks across different views is constructed, where each connected component of the graph corresponds to a distinct part. The process is illustrated in \cref{fig:reprojection}.

To ensure cross-state consistency, we first obtain 2D pixel correspondences between images at two states using LoFTR \cite{sun2021loftr}. These matches allow us to map segmentation categories between states, aligning part segmentations across them.

\section{Training Details of GaussianArt}
\label{sec:11}
\noindent{\textbf{Training Implementations}.} The number of initialized Gaussians is 5,000.
During the warm-up period, we train the Gaussians for 6,000 iterations, supervised by RGB-D and part segmentation masks at canonical states, initializing Gaussians' color and geometry while regularizing the weights. The process takes about 2 minutes. During this process, we use the densification strategy of 3DGS and deactivate it during motion learning. 

Subsequently, we set up 4,000 steps of soft-training under the supervision of RGB-D and part segmentation masks from two states, rigidifying the parts and learning basic motions. This stage modifies Gaussians that were incorrectly initialized across different parts, allowing the positions and weights of the Gaussians to gradually stabilize during the motion learning process. This period takes about 6 minutes.

During hard training, we treat the Gaussians as rigid parts and apply simple and efficient motion estimation to focus on motion learning for each part, resulting in accurate performance. 

For the training process, $\lambda_{\text{SSIM}}$ is 0.2, $\lambda_{\text{D}}$ is 0.5, $\lambda_{\text{SEM}}$ is 0.5, $\lambda_{\text{sparsity}}$ is 1.0, and $\lambda_{\text{traj}}$ is 1.0. The rotation degree threshold $\epsilon$ is set to 15°. The learning rate $r$ of Gaussians' positions changes with motion learning as follows:

\begin{equation}
    \text{lr}(r) =
    \begin{cases}
    \text{max\_lr} \cdot \left( \frac{\text{min\_lr}}{\text{max\_lr}} \right)^{\frac{r- \text{init}}{\text{end}-\text{init}}}, & \text{init} \leq r < \text{end} \\
    \text{min\_lr}, & r \geq \text{end}
    \end{cases}
\end{equation}
\label{eq:17}
\\
where "max\_lr" is $1.6\times10^{-4}$, "min\_lr" is $1.0\times10^{-8}$, "init" is 6000, and "end" is 10000.

Moreover, for mesh extraction, the voxel size is 0.005 and the truncated threshold is 0.04, with space carving.

\noindent{\textbf{Correspondence Filtering}.} We introduce a trajectory regularization term to provide extra supervision on motion parameters. We further use a 3D locality filter to refine the matching results. In this section, we detail the calculation of the filter. 

Given matching pixel pairs from two views at two states $\{\mathbf{p}_i, \mathbf{q}_i\}_{i=1}^{N_m}$, we project them as 3D points $\{\tilde{\mathbf{p}}_i, \tilde{\mathbf{q}}_i\}_{i=1}^{N_m}$ with the depth map and the camera intrinsics and extrinsics:

\begin{equation}
    \tilde{\mathbf{p}}_i = \mathbf{D}_1(\mathbf{p}_i)\odot(\mathbf{K}_1^{-1}{\mathbf{E}_1}),
    \label{eq:18}
\end{equation}
\begin{equation}
 \tilde{\mathbf{q}}_i = \mathbf{D}_2(\mathbf{q}_i)\odot(\mathbf{K}_2^{-1}{\mathbf{E}_2}).
 \label{eq:19}
\end{equation}
\\
However, the matching results include false correspondences, which may adversely affect the regularization process. To address this issue, we apply a 3D locality filter to enhance the accuracy of the correspondences. For each starting-state point in the matching pair $\mathbf{p}_i$, we first query adjacent points to form a neighborhood set in the starting state:

\begin{equation}
    \mathcal{N}(\tilde{\mathbf{p}}_i) = \left\{\tilde{\mathbf{p}}_j: \Vert\tilde{\mathbf{p}}_i - \tilde{\mathbf{p}}_j\Vert < r\right\}.
    \label{eq:20}
\end{equation}
\\
The ending state neighborhood set can be formulated similarly:

\begin{equation}
    \mathcal{N}^\prime(\tilde{\mathbf{q}}_i) = \left\{\tilde{\mathbf{q}}_j: \Vert\tilde{\mathbf{q}}_i - \tilde{\mathbf{q}}_j\Vert < r\right\},
    \label{eq:21}
\end{equation}
\\
where $\tilde{\mathbf{q}}_i$ is the corresponding point to $\tilde{\mathbf{p}}_i$ in the matching pair.

The local geometric structure is invariant to rigid transformations. Therefore, a key characteristic of false matches is their significant deviation from the transformed set center. Based on this observation, we define the 3D locality filter as:

\begin{equation}
    F = \left\{\begin{array}{l}
        1 \textnormal{ if } \Vert\tilde{\mathbf{q}}_j - \mathbf{m}_q\Vert < r^\prime, \\
        0 \textnormal{ else },
    \end{array}\right.
    \label{eq:22}
\end{equation}
\\
where $\mathbf{m}_q = \frac{1}{\left\vert\mathcal{N}^\prime(\mathbf{p}_i)\right\vert}\sum_{j\in\mathcal{N}^\prime(\mathbf{p}_i)}\tilde{\mathbf{q}}_j$ is the mean of the ending state point set. We set the two thresholds to $r = 0.01$ and $r^\prime = 0.02$ separately.

\section{Mesh Extraction}
\label{sec:12}
To extract meshes from Gaussians, we employ the method of rendering median depth as introduced in \cite{keetha2024splatam} and fuse it into a mesh using VDBFusion \cite{vizzo2022vdbfusion}. This entire process can be efficiently accomplished with GauS \cite{ye2024gaustudio} with appropriate selections of voxel size and truncated threshold.

\section{More Discussion of Gaussian-to-Mesh}
\label{sec:13}
Although this work does not technically explore mesh reconstruction from Gaussians, the interesting phenomena observed during the experimental process still inspire us.

For objects with severely uneven view distributions, such as the USB and Stapler in PARIS \cite{liu2023paris}, the mesh method employed in this work fails to complete regions with sparse views, leading to a low CD.
Additionally, objects with noisy depth maps, such as real objects, may result in holes in the extracted mesh. While a tetrahedral grid-based method in \cite{yu2024gaussian} can mitigate these issues, it often introduces surface noise when using vanilla 3DGS. In order to optimize surface reconstruction, we attempted to adapt the method from GaussianArt to Gaussian Opacity Fields \cite{yu2024gaussian} or impose surface constraints on the vanilla baseline. However, none of these approaches could effectively learn the articulated motion. This, to some extent, indicates that the modeling methods for flattened Gaussians in articulated objects require further exploration. In future research, we will explore more effective mesh reconstruction methods while maintaining accurate motion estimation to improve mesh quality under different conditions.

\section{Failure Case}
\label{sec:14}
\begin{figure}[h!]
    \centering
    \includegraphics[width=0.95\linewidth]{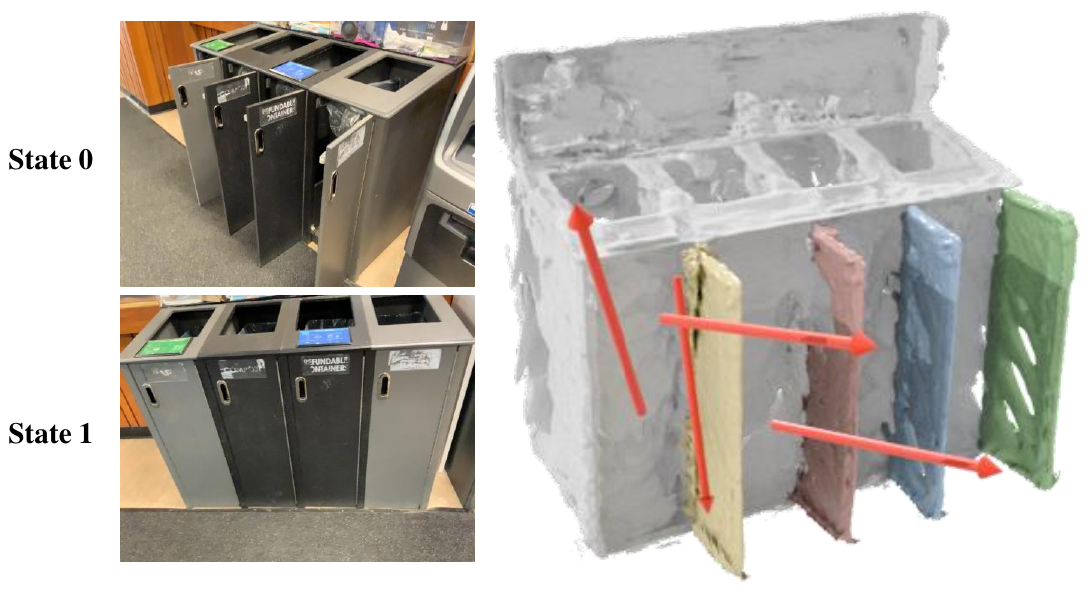}
    \vspace{-3mm}
    \caption{Failure case.}
    \label{fig:fail}
    \vspace{-3mm}
\end{figure}

When dealing with extreme motions, such as the transition from a fully open to a fully closed door, challenges arise in accurately learning motion parameters, despite our method's ability to successfully segment parts (see \cref{fig:fail}). 

In future work, we aim to explore strategies for imposing constraints on intermediate motion states within the unified GS framework to enhance the robustness of motion learning.

\section{Basic Datasets and Metrics}
\label{sec:15}
\subsection{Datasets}
\noindent{\textbf{PARIS Two-Part Dataset.}} The dataset created by \cite{liu2023paris} contains multi-view posed renderings of objects spanning 10 categories in PM, along with 2 real world scans of articulated objects of the same fashion. We follow \cite{weng2024digitaltwin} to use an enhanced version with rendered depth maps. 

\noindent{\textbf{DigitalTwinArt-PM Dataset.}} A multi-part dataset proposed by \cite{weng2024digitaltwin}, containing two 3-part articulated objects from PM, each with one static part and two movable parts. 

\noindent{\textbf{GS-PM Dataset.}} To study articulated objects with more parts, we create GS-PM. The objects we include consist of at most 7 parts and complex combinations of motion types, serving as a significantly stronger benchmark for evaluation.

\subsection{Evaluation Metrics}
\noindent{\textbf{Motion Parameters Estimation}} For the estimated motion, we first interpret unified rigid motion matrices as rotation axes (axis origin and axis direction) and joint states. We then calculate the following metrics: \textbf{Axis Pos Error (0.1m)}, which measures the Euclidean distance between the estimated axis origin and ground truth; \textbf{Axis Angle Error ($^\circ$)}, which measures the angular deviation of the estimated axis direction; \textbf{Part Motion Error} ($^\circ$ for revolute joints and m for prismatic joints) to measure the difference.

\noindent{\textbf{Geometric Reconstruction Quality}} We use CD, calculated on 10,000 uniformly sampled points from the reconstructed meshes and the ground truth meshes. To evaluate the quality of part-level reconstruction, we further measure \textbf{CD-d (mm)} on the dynamic parts, \textbf{CD-s (mm)} on the static part and \textbf{CD-w (mm)} on the whole mesh.

\noindent{\textbf{Visual Reconstruction Quality}}
We also report the average PSNR and SSIM for novel views in all objects. 

\section{Results on Basic Datasets}
\label{sec:16}
\sisetup{detect-all=true}
\begin{table*}[t]
\centering
\def\mywidth{0.99\textwidth} 
\def\myspc{}{\hspace{8ex}}
\resizebox{\mywidth}{!}{
\sisetup{table-auto-round,table-format=.2, table-column-width=1.35cm}
\begin{tabular}{cl|SSSSSSSSSSS[table-column-width=1.6cm]|SSS}

\hline
                               &                                & \multicolumn{11}{c|}{Simulation}                                                                                     & \multicolumn{3}{c}{Real } \bigstrut[t]\\
                               &                                & \multicolumn{1}{c}{FoldChair}                      & \multicolumn{1}{c}{Fridge}                         & \multicolumn{1}{c}{Laptop\textsuperscript{\dag}}   & \multicolumn{1}{c}{Oven\textsuperscript{\dag}}     & \multicolumn{1}{c}{Scissor}                        & \multicolumn{1}{c}{Stapler}                        & \multicolumn{1}{c}{USB}                            & \multicolumn{1}{c}{Washer}                         & \multicolumn{1}{c}{Blade}                          & \multicolumn{1}{c}{Storage\textsuperscript{\dag}}  & \multicolumn{1}{c|}{All}                            & \multicolumn{1}{c}{Fridge}                         & \multicolumn{1}{c}{Storage}                        & \multicolumn{1}{c}{All \bigstrut[b]}\\
\hline

\multirow{4}[2]{*}{\shortstack{Axis\\Ang}} 
& Ditto~\cite{Jiang_2022_ditto}    & 89.35                          & 89.30*                         & 3.12                           & 0.96                           & 4.50                           & 89.86                          & 89.77                          & 89.51                          & 79.54*                         & 6.32                           & 54.22                          & 1.71                           &  5.88                  &  3.80 \bigstrut[t]\\

& PARIS~\cite{liu2023paris}      & 8.08\tsb{13.2}                 & 9.15\tsb{28.3}                 & \cellcolor{tabsecond} 0.02\tsb{0.0}         & 0.04\tsb{0.0}         & 3.82\tsb{3.4}                  & 39.73\tsb{35.1}                & 0.13\tsb{0.2}                  & 25.36\tsb{30.3}                & 15.38\tsb{14.9}                & 0.03\tsb{0.0}         & 10.17\tsb{12.5}                & \cellcolor{tabsecond}1.64\tsb{0.3}         & 43.13\tsb{23.4}                & 22.39\tsb{11.9} \\
                               
& PARIS*~\cite{liu2023paris}     & 15.79\tsb{29.3}                & 2.93\tsb{5.3}                  & 0.03\tsb{0.0}                  & 7.43\tsb{23.4}                 & 16.62\tsb{32.1}                & 8.17\tsb{15.3}                 & 0.71\tsb{0.8}                  & 18.40\tsb{23.3}                & 41.28\tsb{31.4}                & 0.03\tsb{0.0}         & 11.14\tsb{16.1}                & 1.90\tsb{0.0}                  & 30.10\tsb{10.4}                & 16.00\tsb{5.2} \\
                                
& CSG-reg \cite{weng2024digitaltwin}                        & 0.10\tsb{0.0}                  & 0.27\tsb{0.0}                 & 0.47\tsb{0.0}                 & 0.35\tsb{0.1}                 & 0.28\tsb{0.0}                 & 0.30\tsb{0.0}                 & 11.78\tsb{10.5}               & 71.93\tsb{6.3}                 & 7.64\tsb{5.0}                 & 2.82\tsb{2.5}                 & 9.60\tsb{2.4}                 & 8.92\tsb{0.9}                 & 69.71\tsb{9.6}                & 39.31\tsb{5.2} \\

& 3Dseg-reg \cite{weng2024digitaltwin}                      &\multicolumn{1}{c}{-}                             &\multicolumn{1}{c}{-}                             & 2.34\tsb{0.11}                 &\multicolumn{1}{c}{-}                             &\multicolumn{1}{c}{-}                             &\multicolumn{1}{c}{-}                             &\multicolumn{1}{c}{-}                             &\multicolumn{1}{c}{-}                             & 9.40\tsb{7.5}                 &\multicolumn{1}{c}{-}                             &\multicolumn{1}{c}{-}                             &\multicolumn{1}{c}{-}                             &\multicolumn{1}{c}{-}                             &\multicolumn{1}{c}{-}\\
                         
& DigitalTwinArt \cite{weng2024digitaltwin} & 0.03\tsb{0.0}& 0.07\tsb{0.0} & 0.06\tsb{0.0} & 0.22\tsb{0.0} &  0.11\tsb{0.0} & 0.06\tsb{0.0} & 0.11\tsb{0.0} & 0.43\tsb{0.0} & 0.27\tsb{0.0} & 0.06\tsb{0.0} & 0.14\tsb{0.0} & 2.10\tsb{0.0} & 18.11\tsb{0.2} & 10.11\tsb{0.1} \bigstrut[b]\\

& ArtGS \cite{liu2025artgs} & \bfseries\cellcolor{tabfirst} 0.01\tsb{0.0}& \bfseries \cellcolor{tabfirst}0.03\tsb{0.0} &\bfseries\cellcolor{tabfirst} 0.01\tsb{0.0} &\bfseries\cellcolor{tabfirst} 0.01\tsb{0.0} & \cellcolor{tabsecond} 0.05\tsb{0.0} & \bfseries\cellcolor{tabfirst} 0.01\tsb{0.0} & \cellcolor{tabsecond} 0.04\tsb{0.0} & \bfseries\cellcolor{tabfirst} 0.02\tsb{0.0} & \cellcolor{tabsecond} 0.03\tsb{0.0} &\bfseries\cellcolor{tabfirst} 0.01\tsb{0.0} & \bfseries\cellcolor{tabfirst} 0.02\tsb{0.0} & 2.09\tsb{0.0} &\cellcolor{tabsecond} 3.47\tsb{0.3} &\cellcolor{tabsecond} 2.78\tsb{0.2} \bigstrut[b]\\

& Ours & \cellcolor{tabsecond} 0.02\tsb{0.0}& \bfseries \cellcolor{tabfirst} 0.03\tsb{0.0} &\cellcolor{tabsecond}0.02\tsb{0.0} &\bfseries   \cellcolor{tabfirst} 0.01\tsb{0.0} & \bfseries  \cellcolor{tabfirst} 0.04\tsb{0.0} & \bfseries \cellcolor{tabsecond} 0.02\tsb{0.0} & \bfseries \cellcolor{tabfirst} 0.01\tsb{0.0} & \cellcolor{tabsecond} 0.04\tsb{0.0} & \bfseries \cellcolor{tabfirst} 0.01\tsb{0.0} & \bfseries  \cellcolor{tabfirst} 0.01\tsb{0.0} & \bfseries \cellcolor{tabfirst}0.02\tsb{0.0} & \bfseries \cellcolor{tabfirst}1.38\tsb{0.1} &\bfseries \cellcolor{tabfirst} 3.07\tsb{0.2} & \bfseries\cellcolor{tabfirst}2.23\tsb{0.2} \bigstrut[b]\\
\hline

\multirow{4}[2]{*}{\shortstack{Axis\\Pos}} & Ditto~\cite{Jiang_2022_ditto}    & 3.77                           & 1.02*                          & 0.01                           & 0.13                           & 5.70                           & 0.20                           & 5.41                           & 0.66                           &\multicolumn{1}{c}{-}                             &\multicolumn{1}{c}{-}                             & 2.11                           & 1.84                           &\multicolumn{1}{c}{-}                             & 1.84 \bigstrut[t]\\

& PARIS~\cite{liu2023paris}      & 0.45\tsb{0.9}                  & 0.38\tsb{1.0}                  &  \bfseries \cellcolor{tabfirst}0.00\tsb{0.0}         &  \bfseries \cellcolor{tabfirst}0.00\tsb{0.0}         & 2.10\tsb{1.4}                  & 2.27\tsb{3.4}                  & 2.36\tsb{3.4}                  & 1.50\tsb{1.3}                  &\multicolumn{1}{c}{-}                             &\multicolumn{1}{c}{-}                             & 1.13\tsb{1.1}                  & \bfseries \cellcolor{tabfirst} 0.34\tsb{0.2}         &\multicolumn{1}{c}{-}                             & \bfseries \cellcolor{tabfirst}0.34\tsb{0.2} \\
                               
& PARIS*~\cite{liu2023paris}     & 0.25\tsb{0.5}                  & 1.13\tsb{2.6}                  &  \bfseries \cellcolor{tabfirst} 0.00\tsb{0.0}         & 0.05\tsb{0.2}                  & 1.59\tsb{1.7}                  & 4.67\tsb{3.9}                  & 3.35\tsb{3.1}                  & 3.28\tsb{3.1}                  &\multicolumn{1}{c}{-}                             &\multicolumn{1}{c}{-}                             & 1.79\tsb{1.5}                  & 0.50\tsb{0.0}                  &\multicolumn{1}{c}{-}                             & 0.50\tsb{0.0} \\
                                
& CSG-reg \cite{weng2024digitaltwin}                        & 0.02\tsb{0.0}                  & \bfseries \cellcolor{tabfirst}0.00\tsb{0.0}         & 0.20\tsb{0.2}                 & 0.18\tsb{0.0}                 & 0.01\tsb{0.0}        & 0.02\tsb{0.0}                  & 0.01\tsb{0.0}                 & 2.13\tsb{1.5}                 &\multicolumn{1}{c}{-}                             &\multicolumn{1}{c}{-}                             & 0.32\tsb{0.2}                 & 1.46\tsb{1.1}                 &\multicolumn{1}{c}{-}                             & 1.46\tsb{1.1} \\
                               
& 3Dseg-reg \cite{weng2024digitaltwin}                      &\multicolumn{1}{c}{-}                             &\multicolumn{1}{c}{-}                             & 0.10\tsb{0.0}                 &\multicolumn{1}{c}{-}                             &\multicolumn{1}{c}{-}                             &\multicolumn{1}{c}{-}                             &\multicolumn{1}{c}{-}                             &\multicolumn{1}{c}{-}                             &\multicolumn{1}{c}{-}                             &\multicolumn{1}{c}{-}                             &\multicolumn{1}{c}{-}                             &\multicolumn{1}{c}{-}                             &\multicolumn{1}{c}{-}                             &\multicolumn{1}{c}{-}\\                              

& DigitalTwinArt \cite{weng2024digitaltwin} & 0.01\tsb{0.0} & 0.01\tsb{0.0} & \bfseries\cellcolor{tabfirst} 0.00\tsb{0.0} & 0.01\tsb{0.0} & 0.02\tsb{0.0} & 0.01\tsb{0.0} & \bfseries\cellcolor{tabfirst} 0.00\tsb{0.0} &\cellcolor{tabsecond}0.01\tsb{0.0} &  \multicolumn{1}{c}{-} & \multicolumn{1}{c}{-} & 0.01\tsb{0.0} & 0.57\tsb{0.0} & \multicolumn{1}{c}{-} & 0.57\tsb{0.0} \bigstrut[b]\\

& ArtGS \cite{liu2025artgs} & \bfseries\cellcolor{tabfirst} 0.00\tsb{0.0}& \bfseries \cellcolor{tabfirst}0.00\tsb{0.0} & 0.01\tsb{0.0} & \bfseries \cellcolor{tabfirst}0.00\tsb{0.0} & \bfseries \cellcolor{tabfirst} 0.00\tsb{0.0} & 0.01\tsb{0.0} & \bfseries \cellcolor{tabfirst} 0.00\tsb{0.0} &\bfseries \cellcolor{tabfirst} 0.00\tsb{0.0} & \multicolumn{1}{c}{-} &  \multicolumn{1}{c}{-}  & \bfseries \bfseries \cellcolor{tabfirst}0.00\tsb{0.0} & 0.47\tsb{0.0} & \multicolumn{1}{c}{-} & 0.47\tsb{0.0} \bigstrut[b]\\

& Ours & \bfseries \cellcolor{tabfirst}0\tsb{0.0} & \bfseries \cellcolor{tabfirst}0\tsb{0.0} & \bfseries\cellcolor{tabfirst} 0\tsb{0.0} & 0.01\tsb{0.0} & \bfseries \cellcolor{tabfirst} 0\tsb{0.0} &  \bfseries \cellcolor{tabfirst} 0\tsb{0.0} &  \bfseries \cellcolor{tabfirst} 0\tsb{0.0} &\cellcolor{tabsecond}0.01\tsb{0.0} &  \multicolumn{1}{c}{-} & \multicolumn{1}{c}{-} &  \bfseries \cellcolor{tabfirst} 0\tsb{0.0} & \cellcolor{tabsecond}0.4\tsb{0.0} & \multicolumn{1}{c}{-} &\cellcolor{tabsecond} 0.4\tsb{0.0} \bigstrut[b]\\
\hline

\multirow{4}[2]{*}{\shortstack{Part\\Motion}} 
& Ditto~\cite{Jiang_2022_ditto}    & 99.36                          & F                              & 5.18                           & 2.09                           & 19.28                          & 56.61                          & 80.60                          & 55.72                          & F                              & 0.09                           & 39.87                          & 8.43                           & 0.38                           & 4.41 \bigstrut[t]\\
                               
& PARIS~\cite{liu2023paris}      & 131.66\tsb{78.9}               & 24.58\tsb{57.7}                & \cellcolor{tabsecond} 0.03\tsb{0.0}         & 0.03\tsb{0.0}         & 120.70\tsb{50.1}               & 110.80\tsb{47.1}               & 64.85\tsb{84.3}                & 60.35\tsb{23.3}                & 0.34\tsb{0.2}                  & 0.30\tsb{0.0}                  & 51.36\tsb{34.2}                & 2.16\tsb{1.1}                  & 0.56\tsb{0.4}                  & 1.36\tsb{0.7} \\
                               
& PARIS*~\cite{liu2023paris}     & 127.34\tsb{75.0}               & 45.26\tsb{58.5}                & \cellcolor{tabsecond} 0.03\tsb{0.0}         & 9.13\tsb{28.8}                 & 68.36\tsb{64.8}                & 107.76\tsb{68.1}               & 96.93\tsb{67.8}                & 49.77\tsb{26.5}                & 0.36\tsb{0.2}                  & 0.30\tsb{0.0}                  & 50.52\tsb{39.0}                & \bfseries \cellcolor{tabfirst} 1.58\tsb{0.0}         & 0.57\tsb{0.1}                  & 1.07\tsb{0.1} \\
                               
& CSG-reg \cite{weng2024digitaltwin}   &0.13\tsb{0.0}         & 0.29\tsb{0.0}                 & 0.35\tsb{0.0}                 & 0.58\tsb{0.0}                 & 0.20\tsb{0.0}                 & 0.44\tsb{0.0}                 & 10.48\tsb{9.3}                & 158.99\tsb{8.8}               & 0.05\tsb{0.0}                 & 0.04\tsb{0.0}                 & 17.16\tsb{1.8}                & 14.82\tsb{0.1}                & 0.64\tsb{0.1}                 & 7.73\tsb{0.1} \\
                               
& 3Dseg-reg \cite{weng2024digitaltwin}                      &\multicolumn{1}{c}{-}                             &\multicolumn{1}{c}{-}                             & 1.61\tsb{0.1}                 &\multicolumn{1}{c}{-}                             &\multicolumn{1}{c}{-}                             &\multicolumn{1}{c}{-}                             &\multicolumn{1}{c}{-}                             &\multicolumn{1}{c}{-}                             & 0.15\tsb{0.0}                 &\multicolumn{1}{c}{-}                             &\multicolumn{1}{c}{-}                             &\multicolumn{1}{c}{-}                             &\multicolumn{1}{c}{-}                             &\multicolumn{1}{c}{-}\\

& DigitalTwinArt \cite{weng2024digitaltwin} & 0.14\tsb{0.0} &0.13\tsb{0.0} & 0.10\tsb{0.0} & 0.15\tsb{0.0} &  0.23\tsb{0.2} & 0.05\tsb{0.0} & 0.11\tsb{0.0} &  0.24\tsb{0.1} & \bfseries \cellcolor{tabfirst} 0.00\tsb{0.0} & \bfseries \cellcolor{tabfirst}0.00\tsb{0.0} & 0.12\tsb{0.0} &  1.79\tsb{0.0} &  0.17\tsb{0.0} & \cellcolor{tabsecond}  0.98\tsb{0.0} \bigstrut[b]\\

& ArtGS \cite{liu2025artgs} & \bfseries\cellcolor{tabfirst} 0.03\tsb{0.0}&  \cellcolor{tabsecond}  0.04\tsb{0.0} &\bfseries\cellcolor{tabfirst} 0.02\tsb{0.0} & \cellcolor{tabsecond}0.02\tsb{0.0} & \bfseries\cellcolor{tabfirst} 0.04\tsb{0.0} & \bfseries\cellcolor{tabfirst} 0.01\tsb{0.0} & \bfseries\cellcolor{tabfirst} 0.03\tsb{0.0} & \bfseries\cellcolor{tabfirst}0.03\tsb{0.0} & \bfseries \cellcolor{tabfirst} 0.00\tsb{0.0} & \bfseries \cellcolor{tabfirst} 0.00\tsb{0.0}\tsb{0.0}  & \bfseries \cellcolor{tabfirst} 0.02\tsb{0.0} & 1.94\tsb{0.0} &\bfseries \cellcolor{tabfirst} 0.04\tsb{0.0} & 0.99\tsb{0.0} \bigstrut[b]\\

& Ours &\bfseries \cellcolor{tabfirst} 0.03\tsb{0.0} 
& \bfseries \cellcolor{tabfirst} 0.02\tsb{0.0} 
&  0.04\tsb{0.0} 
& \bfseries \cellcolor{tabfirst} 0.01\tsb{0.0} 
& \bfseries \cellcolor{tabfirst}0.04\tsb{0.0} 
& \cellcolor{tabsecond} 0.02\tsb{0.0} 
& \bfseries \cellcolor{tabfirst} 0.03\tsb{0.0} 
& \cellcolor{tabsecond} 0.06\tsb{0.0} 
& \bfseries \cellcolor{tabfirst} 0\tsb{0.0} 
& \bfseries \cellcolor{tabfirst} 0\tsb{0.0} 
& \cellcolor{tabsecond} 0.03\tsb{0.0} 
& \cellcolor{tabsecond} 1.68\tsb{0.0} 
& \cellcolor{tabsecond}  0.15\tsb{0.0}
& \bfseries \cellcolor{tabfirst} 0.92\tsb{0.0} \bigstrut[b]\\
\hline
\multirow{4}[2]{*}{CD-s}       
& Ditto~\cite{Jiang_2022_ditto}    & 33.79                          & 3.05                           & 0.25                           
& \bfseries \cellcolor{tabfirst}2.52                  & 39.07                          & 41.64                          & 2.64                           & 10.32                          & 46.9                           & 9.18                           & 18.94                          & 47.01                          & 16.09                          & 31.55 \bigstrut[t]\\
                               
& PARIS~\cite{liu2023paris}      & 9.16\tsb{5.0}                  
& 3.65\tsb{2.7}                  
& \bfseries \cellcolor{tabfirst} 0.16\tsb{0.0}         
& 12.95\tsb{1.0}                 & 1.94\tsb{3.8}                  
& \bfseries \cellcolor{tabfirst}1.88\tsb{0.2}         & 2.69\tsb{0.3}                  & 25.39\tsb{2.2}                 & 1.19\tsb{0.6}                  & 12.76\tsb{2.5}                 & 7.18\tsb{1.8}                  & 42.57\tsb{34.1}                & 54.54\tsb{30.1}                & 48.56\tsb{32.1} \\
                               
& PARIS*~\cite{liu2023paris}     & 10.20\tsb{5.8}                 
& 8.82\tsb{12.0}                 
& \bfseries \cellcolor{tabfirst}0.16\tsb{0.0}         
& 3.18\tsb{0.3}                  
& 15.58\tsb{13.3}               
& 2.48\tsb{1.9}                  
& \cellcolor{tabsecond} 1.95\tsb{0.5}         
& 12.19\tsb{3.7}                 & 1.40\tsb{0.7}                  & 8.67\tsb{0.8}                  & 6.46\tsb{3.9}                  & 11.64\tsb{1.5}                 & 20.25\tsb{2.8}                 & 15.94\tsb{2.1} \\
                               
& CSG-reg \cite{weng2024digitaltwin}                        & 1.69                  & 1.45                  & 0.32         & 3.93         & 3.26                  & \cellcolor{tabsecond}2.22         &  \cellcolor{tabsecond}1.95         & \bfseries \cellcolor{tabfirst} 4.53         & 0.59                  & 7.06                  & 2.70                  & 6.33                  & 12.55                 & 9.44 \\
                               
& 3Dseg-reg \cite{weng2024digitaltwin}                      &\multicolumn{1}{c}{-}                             &\multicolumn{1}{c}{-}                             & 0.76                  &\multicolumn{1}{c}{-}                             &\multicolumn{1}{c}{-}                             &\multicolumn{1}{c}{-}                             &\multicolumn{1}{c}{-}                             &\multicolumn{1}{c}{-}                             & 66.31                 &\multicolumn{1}{c}{-}                             &\multicolumn{1}{c}{-}                             &\multicolumn{1}{c}{-}                             &\multicolumn{1}{c}{-}                             &\multicolumn{1}{c}{-}\\

& DigitalTwinArt \cite{weng2024digitaltwin} & \cellcolor{tabsecond} 0.18\tsb{0.0} & 0.60\tsb{0.0} & 0.31\tsb{0.0} & 4.55\tsb{0.1} & \bfseries \cellcolor{tabfirst} 0.39\tsb{0.0} & 2.85\tsb{0.1} & 2.10\tsb{0.0} & \cellcolor{tabsecond}5.02\tsb{0.2} & \cellcolor{tabsecond} 0.44\tsb{0.0} & \cellcolor{tabsecond} 4.95\tsb{0.2} & \cellcolor{tabsecond} 2.14\tsb{0.1} & 2.74\tsb{0.2} & 9.53\tsb{0.3} &  6.14\tsb{0.3} \bigstrut[b]\\

& ArtGS \cite{liu2025artgs} & 0.26\tsb{0.3}& \bfseries \cellcolor{tabfirst} 0.52\tsb{0.0} & 0.63\tsb{0.0} & 3.88\tsb{0.0} & 0.61\tsb{0.3} & 3.83\tsb{0.1} & 2.25\tsb{0.2} & 6.43\tsb{0.1} & 0.54\tsb{0.0} & 7.31\tsb{0.2}  & 2.63\tsb{0.1} &\cellcolor{tabsecond}  1.64\tsb{0.2} &\cellcolor{tabsecond} 2.93\tsb{0.3} &\cellcolor{tabsecond} 2.29\tsb{0.3} \bigstrut[b]\\

& Ours 
& \bfseries \cellcolor{tabfirst} 0.13\tsb{0.0} 
& \bfseries \cellcolor{tabfirst} 0.52\tsb{0.0} 
& 0.20\tsb{0.0} 
& \cellcolor{tabsecond}2.60\tsb{0.1} 
& \cellcolor{tabsecond} 0.40\tsb{0.0} 
& 3.25\tsb{0.1} 
& \bfseries \cellcolor{tabfirst}1.94\tsb{0.1}
& 5.39\tsb{0.2} 
& \bfseries \cellcolor{tabfirst}0.43\tsb{0.0} 
& \bfseries \cellcolor{tabfirst}3.60\tsb{0.1} 
& \bfseries \cellcolor{tabfirst}1.85\tsb{0.1} 
& \bfseries \cellcolor{tabfirst}1.32\tsb{0.1} 
& \bfseries \cellcolor{tabfirst}2.92\tsb{0.1} 
& \bfseries \cellcolor{tabfirst}2.12\tsb{0.1} \bigstrut[b]\\

\hline
\multirow{4}[2]{*}{CD-m}       
& Ditto~\cite{Jiang_2022_ditto}    & 141.11                         & 0.99                           & 0.19                           & 0.94                           & 20.68                          & 31.21                          & 15.88                          & 12.89                          & 195.93                         & 2.20                           & 42.20                          & 50.60                          &   20.35                 & 35.48 \bigstrut[t]\\
                               
& PARIS~\cite{liu2023paris}      & 8.99\tsb{7.6}                  & 7.76\tsb{11.2}                 & 0.21\tsb{0.2}                  & 28.70\tsb{15.2}                & 46.64\tsb{40.7}                & 19.27\tsb{30.7}                & 5.32\tsb{5.9}                  & 178.43\tsb{131.7}              & 25.21\tsb{9.5}                 & 76.69\tsb{6.1}                 & 39.72\tsb{25.9}                & 45.66\tsb{31.7}                & 864.82\tsb{382.9}              & 455.24\tsb{207.3} \\
                               
& PARIS*~\cite{liu2023paris}     & 17.97\tsb{24.9}                & 7.23\tsb{11.5}                 & \cellcolor{tabsecond} 0.15\tsb{0.0}         & 6.54\tsb{10.6}                 & 16.65\tsb{16.6}                & 30.46\tsb{37.0}                & 10.17\tsb{6.9}                 & 265.27\tsb{248.7}              & 117.99\tsb{213.0}              & 52.34\tsb{11.0}                & 52.48\tsb{58.0}                & 77.85\tsb{26.8}                & 474.57\tsb{227.2}              & 276.21\tsb{127.0} \\
                                
& CSG-reg \cite{weng2024digitaltwin}                        & 1.91                  & 21.71                 & 0.42                  & 256.99                & 1.95                  & 6.36                  & 29.78                 & 436.42                & 26.62                 & 1.39                  & 78.36                 & 442.17                & 521.49                & 481.83 \\
                               
& 3Dseg-reg \cite{weng2024digitaltwin}                      &\multicolumn{1}{c}{-}                             &\multicolumn{1}{c}{-}                             & 1.01                  &\multicolumn{1}{c}{-}                             &\multicolumn{1}{c}{-}                             &\multicolumn{1}{c}{-}                             &\multicolumn{1}{c}{-}                             &\multicolumn{1}{c}{-}                             & 6.23                  &\multicolumn{1}{c}{-}                             &\multicolumn{1}{c}{-}                             &\multicolumn{1}{c}{-}                             &\multicolumn{1}{c}{-}                             &\multicolumn{1}{c}{-}\\
                          
& DigitalTwinArt \cite{weng2024digitaltwin} 
& \bfseries \cellcolor{tabfirst} 0.15\tsb{0.0} 
& 0.27\tsb{0.0} 
& 0.16\tsb{0.0} 
& \bfseries \cellcolor{tabfirst}  0.44\tsb{0.0} 
& \bfseries \cellcolor{tabfirst}0.37\tsb{0.0} 
& 1.45\tsb{0.4} 
&  1.54\tsb{0.2} 
& \cellcolor{tabsecond} 0.30\tsb{0.0}
&  1.73\tsb{0.1} 
& \bfseries \cellcolor{tabfirst} 0.40\tsb{0.0} 
& 0.68\tsb{0.1} 
& 1.24\tsb{0.0} 
& 28.58\tsb{5.0}
& 14.91\tsb{2.5} \bigstrut[b]\\

& ArtGS \cite{liu2025artgs} & 0.54\tsb{0.1}&\cellcolor{tabsecond} 0.21\tsb{0.0} & \bfseries \cellcolor{tabfirst} 0.13\tsb{0.0} &\cellcolor{tabsecond} 0.89\tsb{0.2} & 0.64\tsb{0.4} &  \cellcolor{tabsecond} 0.52\tsb{0.1} &\bfseries \cellcolor{tabfirst} 1.22\tsb{0.1} & 0.45\tsb{0.2} & \cellcolor{tabsecond} 1.12\tsb{0.2} &\cellcolor{tabsecond} 1.02\tsb{0.4}  & \cellcolor{tabsecond} 0.67\tsb{0.2} &\bfseries \cellcolor{tabfirst} 0.66\tsb{0.2} &\cellcolor{tabsecond} 6.28\tsb{3.6} &\cellcolor{tabsecond} 3.47\tsb{1.9} \bigstrut[b]\\

& Ours 
& \cellcolor{tabsecond} 0.29\tsb{0.0}
& \bfseries \cellcolor{tabfirst} 0.18\tsb{0.0} 
& 0.16\tsb{0.0} 
& 1.47\tsb{0.0} 
& \cellcolor{tabsecond}0.39\tsb{0.0} 
& \bfseries \cellcolor{tabfirst} 0.46\tsb{0.0} 
& \cellcolor{tabsecond} 1.31\tsb{0.0} 
& \bfseries \cellcolor{tabfirst} 0.25\tsb{0.0} 
& \bfseries \cellcolor{tabfirst} 0.5\tsb{0.0} 
& 1.08\tsb{0.1}
& \bfseries \cellcolor{tabfirst} 0.61\tsb{0.0} 
&\cellcolor{tabsecond} 0.90\tsb{0.0} 
& \bfseries \cellcolor{tabfirst}3.93\tsb{0.0} 
& \bfseries \cellcolor{tabfirst}2.42\tsb{0.0} \bigstrut[b]\\
\hline
\multirow{4}[2]{*}{CD-w}       
& Ditto~\cite{Jiang_2022_ditto}    & 6.80                           & 2.16                           & 0.31                           
& \bfseries \cellcolor{tabfirst} 2.51                  & 1.70                           & 2.38                           & 2.09                           & 7.29                           & 42.04                          
&  \cellcolor{tabsecond}3.91                  & 7.12                           & 6.50                           & 14.08                          & 10.29 \bigstrut[t]\\
                               
& PARIS~\cite{liu2023paris}      & 1.80\tsb{1.2}                  & 2.92\tsb{0.9}                  & \cellcolor{tabsecond}0.30\tsb{0.1}                  & 11.73\tsb{1.1}                 & 10.49\tsb{20.7}                & 3.58\tsb{4.2}                  & 2.00\tsb{0.2}                  & 24.38\tsb{3.3}                 & 0.60\tsb{0.2}                  & 8.57\tsb{0.4}                  & 6.64\tsb{3.2}                  & 22.98\tsb{15.5}                & 63.35\tsb{22.2}                & 43.16\tsb{18.9} \\
                               
& PARIS*~\cite{liu2023paris}     & 4.37\tsb{6.4}                  & 5.53\tsb{4.7}                  & \bfseries \cellcolor{tabfirst} 0.26\tsb{0.0}         & 3.18\tsb{0.3}                  & 3.90\tsb{3.6}                  & 5.27\tsb{5.9}                  & 1.78\tsb{0.2}                  & 10.11\tsb{2.8}                 & 0.58\tsb{0.1}                  & 7.80\tsb{0.4}                  & 4.28\tsb{2.4}                  & 8.99\tsb{1.4}                  & 32.10\tsb{8.2}                 & 20.55\tsb{4.8} \\
                                
& CSG-reg \cite{weng2024digitaltwin}                        & 0.48                  & 0.98                  & 0.40                  & 3.00         & 1.70                  
& \cellcolor{tabsecond}1.99         & \cellcolor{tabsecond} 1.20                  
& \cellcolor{tabsecond}4.48         & 0.56                  & 4.00                  & 1.88                  & 5.71                  & 14.29                 & 10.00 \\
                               
& 3Dseg-reg \cite{weng2024digitaltwin}                      &\multicolumn{1}{c}{-}                             &\multicolumn{1}{c}{-}                             & 0.81                  &\multicolumn{1}{c}{-}                             &\multicolumn{1}{c}{-}                             &\multicolumn{1}{c}{-}                             &\multicolumn{1}{c}{-}                             &\multicolumn{1}{c}{-}                             & 0.78                  &\multicolumn{1}{c}{-}                             &\multicolumn{1}{c}{-}                             &\multicolumn{1}{c}{-}                             &\multicolumn{1}{c}{-}                             &\multicolumn{1}{c}{-}\\
                             
& DigitalTwinArt \cite{weng2024digitaltwin} 
& \bfseries \cellcolor{tabfirst}  0.27\tsb{0.0}
& 0.70\tsb{0.0} 
& 0.33\tsb{0.0} 
& 4.14\tsb{0.1} 
& \bfseries \cellcolor{tabfirst}0.40\tsb{0.0}
& \bfseries \cellcolor{tabfirst}1.92\tsb{0.1} 
&  1.28\tsb{0.2} 
& \bfseries \cellcolor{tabfirst} 4.36\tsb{0.2} 
& \bfseries \cellcolor{tabfirst}0.36\tsb{0.0} 
& 3.97\tsb{0.2} 
& \cellcolor{tabsecond} 1.77\tsb{0.1} 
& 2.20\tsb{0.1}
& 8.03\tsb{0.5}
& 5.12\tsb{0.3} \bigstrut[b]\\

& ArtGS \cite{liu2025artgs} & 0.43\tsb{0.2}& \bfseries\cellcolor{tabfirst} 0.58\tsb{0.0} & 0.50\tsb{0.0} & 3.58\tsb{0.0} & 0.67\tsb{0.3} & 2.63\tsb{0.0} & 1.28\tsb{0.0} & 5.99\tsb{0.1} & 0.61\tsb{0.0} & 5.21\tsb{0.1}  & 2.15\tsb{0.1} &\bfseries \cellcolor{tabfirst} 1.29\tsb{0.1} &\bfseries \cellcolor{tabfirst} 3.23\tsb{0.1} &\bfseries \cellcolor{tabfirst} 2.26\tsb{0.1} \bigstrut[b]\\

& Ours 
&  \cellcolor{tabsecond}   0.29\tsb{0.0} 
& \bfseries \cellcolor{tabfirst}  0.58\tsb{0.0} 
& 0.47\tsb{0.0}
&  \cellcolor{tabsecond}2.65\tsb{0.1}
& \cellcolor{tabsecond} 0.43\tsb{0.0} 
& 2.35\tsb{0.1}
& \bfseries \cellcolor{tabfirst}1.12\tsb{0.0} 
& 5.03\tsb{0.1} 
& \bfseries \cellcolor{tabfirst} 0.36\tsb{0.0}
& \bfseries \cellcolor{tabfirst}3.6\tsb{0.1} 
& \bfseries \cellcolor{tabfirst}1.69\tsb{0.0}
& \cellcolor{tabsecond}1.36\tsb{0.0}
& \cellcolor{tabsecond}3.35\tsb{0.1}
& \cellcolor{tabsecond}2.36\tsb{0.1} \bigstrut[b]\\
\hline

\end{tabular}
}
\vspace{-2mm}
\caption{Quantitative results on GS-PM Dataset. Metrics are shown as the mean $\pm$ std over 10 trials with different random seeds following \cite{weng2024digitaltwin}. The \colorbox{tabfirst}{best} and \colorbox{tabsecond}{second best} results are highlighted. Objects with \dag \ are the seen categories that Ditto~\cite{Jiang_2022_ditto} has been trained on. Ditto sometimes gives wrong motion type predictions, which are noted with F for joint state and * for joint axis or position. Blade, Storage, and Real Storage have prismatic joints, so there is no Axis Pos.
}

\label{tab:two_part}
\end{table*}
\begin{table*}[t]
\centering
\scriptsize
\setlength\tabcolsep{4pt}
\resizebox{.98\textwidth}{!}{
\begin{tabular}{cc|ccccccccccc|ccc}
\toprule
Metric    & Method &\multicolumn{11}{c|}{Simulation} & \multicolumn{3}{c}{Real} \\
& & FoldChair & Fridge & Laptop & Oven & Scissor & Stapler & USB & Washer & Blade & Storage & All & Fridge & Storage & All \\
\midrule
\multirow{2}*{PSNR} & PARIS \cite{liu2023paris} 
& 33.11 & \cellcolor{tabsecond}38.78&\cellcolor{tabsecond} 38.66 & 35.41 & 38.70 &38.59 & 38.24 & \cellcolor{tabsecond}40.18 & 38.56 &37.08 &37.62 &25.29 & \cellcolor{tabfirst}\textbf{27.13} &26.21 \\

& DigitalTwinArt \cite{weng2024digitaltwin}
&\cellcolor{tabsecond}42.79
&32.99
&37.72
&34.73
&\cellcolor{tabsecond}40.41
&38.28
&38.78
&38.71
&\cellcolor{tabsecond}43.16
&35.38
&\cellcolor{tabsecond}38.30
&24.22
&23.11
&23.67
\\

& ArtGS \cite{liu2025artgs}
&34.46
&37.11
&34.09
&\cellcolor{tabsecond}37.06
&38.29
&\cellcolor{tabsecond}39.13
&\cellcolor{tabsecond}39.64
&38.50
&41.16
&\cellcolor{tabsecond}37.24
&37.67
& \cellcolor{tabfirst}\textbf{27.05}
&25.38
&\cellcolor{tabsecond}26.22
\\

~ & Ours  
& \cellcolor{tabfirst}\textbf{48.37}
& \cellcolor{tabfirst}\textbf{41.24}
& \cellcolor{tabfirst}\textbf{38.70}
& \cellcolor{tabfirst}\textbf{41.15}
& \cellcolor{tabfirst}\textbf{44.60}
& \cellcolor{tabfirst}\textbf{45.69}
& \cellcolor{tabfirst}\textbf{46.02}
& \cellcolor{tabfirst}\textbf{44.41}
& \cellcolor{tabfirst}\textbf{47.72}
& \cellcolor{tabfirst}\textbf{43.52}
& \cellcolor{tabfirst}\textbf{44.14}
& \cellcolor{tabsecond}26.43
&\cellcolor{tabsecond}26.35
& \cellcolor{tabfirst}\textbf{26.39}\\

\midrule
\multirow{2}{*}{SSIM} & PARIS \cite{liu2023paris}  
&0.986 &\cellcolor{tabsecond} 0.994 &\cellcolor{tabsecond} 0.990 &  0.981 &0.996 &0.995 &0.992 & 0.991 & 0.996 & \cellcolor{tabfirst}\textbf{0.994} & 0.992 &0.898 & \cellcolor{tabfirst}\textbf{0.953} &0.926\\
& DigitalTwinArt \cite{weng2024digitaltwin}
&0.986
&0.986
&0.989
&0.977
&0.994
&0.991
&0.991
&0.992
&0.997
&0.968
&0.987
&0.890
&0.923
&0.907
\\

& ArtGS \cite{liu2025artgs}
&\cellcolor{tabfirst}\textbf{0.997}
&0.993
&0.988
&\cellcolor{tabfirst}\textbf{0.995}
&\cellcolor{tabfirst}\textbf{0.998}
&\cellcolor{tabfirst}\textbf{0.999}
&\cellcolor{tabfirst}\textbf{0.998}
&\cellcolor{tabsecond}0.995
&\cellcolor{tabfirst}\textbf{0.999}
&\cellcolor{tabsecond}0.992
&\cellcolor{tabfirst}\textbf{0.995}
&\cellcolor{tabsecond}0.939
&0.930
&\cellcolor{tabsecond}0.935
\\

~ & Ours & \cellcolor{tabsecond}0.994
& \cellcolor{tabfirst}\textbf{0.995}
& \cellcolor{tabfirst}\textbf{0.991}
& \cellcolor{tabsecond}0.992
& \cellcolor{tabfirst}\textbf{0.998}
& \cellcolor{tabsecond}0.997
& \cellcolor{tabsecond}0.997
& \cellcolor{tabfirst}\textbf{0.996}
& \cellcolor{tabfirst}\textbf{0.999}
&0.989
& \cellcolor{tabfirst}\textbf{0.995}
& \cellcolor{tabfirst}\textbf{0.940}
&\cellcolor{tabsecond}0.941
& \cellcolor{tabfirst}\textbf{0.941}

\\
\bottomrule
\end{tabular}}
\vspace{-2mm}
\caption{Quantitative results of the rendering quality from novel views on PARIS Two-Part Dataset. The results are presented on the test set of two states. The \colorbox{tabfirst}{best} and the \colorbox{tabsecond}{second best} results are highlighted.}
\vspace{-3mm}
\label{tab:2pr}
\end{table*}






\begin{table*}[t]
\centering
\def\mywidth{0.99\textwidth} 
\resizebox{\mywidth}{!}{
\begin{tabular}{lc|cccccccccc}
\hline
                               &                                & \multicolumn{1}{c}{Axis Ang 0} & \multicolumn{1}{c}{Axis Ang 1} & \multicolumn{1}{c}{Axis Pos 0} & \multicolumn{1}{c}{Axis Pos 1} & \multicolumn{1}{c}{Part Motion 0} & \multicolumn{1}{c}{Part Motion 1} & \multicolumn{1}{c}{CD-s}       & \multicolumn{1}{c}{CD-m 0}     & \multicolumn{1}{c}{CD-m 1}     & \multicolumn{1}{c}{CD-w}      
                               
                               \bigstrut\\
\hline
\multirow{3}[2]{*}{\shortstack{Fridge\\{10489 (3 parts)}}} & PARIS*~\cite{liu2023paris}     & 34.52                          & 15.91                          & 3.60                           & 1.63                           & 86.21                          & 105.86                         & 8.52                           & 526.19                         & 160.86                         & 15.00 \bigstrut[t]\\
  & DigitalTwinArt \cite{weng2024digitaltwin}   & 0.17   & 0.09  &0.02 &\cellcolor{tabfirst}\textbf{0.00} &0.17 &0.11 & \cellcolor{tabsecond}0.63 &0.44 & 0.57  & 0.89 \bigstrut[t]\\

   & ArtGS \cite{liu2025artgs} & \cellcolor{tabfirst}\textbf{0.02}   & \cellcolor{tabfirst}\textbf{0.00}  &\cellcolor{tabfirst}\textbf{0.00}&\cellcolor{tabfirst}\textbf{0.00} &\cellcolor{tabsecond}0.02 &\cellcolor{tabfirst}\textbf{0.03} & \cellcolor{tabfirst}\textbf{0.62} &\cellcolor{tabfirst}\textbf{0.07} & \cellcolor{tabfirst}\textbf{0.18}  & \cellcolor{tabsecond}0.75 \bigstrut[t]\\
                               & Ours 
                                & \cellcolor{tabfirst}\textbf{0.02} & \cellcolor{tabsecond}0.01 & \cellcolor{tabfirst}\textbf{0.00} & \cellcolor{tabfirst}\textbf{0.00} & \cellcolor{tabfirst}\textbf{0.02} & \cellcolor{tabsecond}0.05 & 0.64 & \cellcolor{tabsecond}0.13 &\cellcolor{tabsecond}0.22
                                &\cellcolor{tabfirst}\textbf{0.56} 
                               \bigstrut[b]\\
\hline

\multirow{3}[2]{*}{\shortstack{Storage\\{47254 (3 parts)}}} & PARIS*~\cite{liu2023paris}     & 43.26                          & 26.18                          & 10.42                          & -                              & 79.84                          & 0.64                           & 8.56                           & 128.62                         & 266.71                         & 8.66 
\bigstrut[t]\\
  & DigitalTwinArt \cite{weng2024digitaltwin}   & 0.16  & 0.80  &0.05 & - &0.16 &\cellcolor{tabfirst}\textbf{0.00} &0.87 & 0.23 & 0.34  & 0.99 
  \bigstrut[t]\\
  
    & ArtGS \cite{liu2025artgs}  & \cellcolor{tabfirst}\textbf{0.01}   &\cellcolor{tabfirst}\textbf{0.02}  &\cellcolor{tabfirst}\textbf{0.01} & - &\cellcolor{tabsecond}0.01 &\cellcolor{tabfirst}\textbf{0.00} &\cellcolor{tabsecond}0.78 & \cellcolor{tabsecond}0.19 & \cellcolor{tabsecond}0.27  & \cellcolor{tabsecond}0.93 
  \bigstrut[t]\\
                               & Ours & \cellcolor{tabsecond}0.02 & \cellcolor{tabsecond}0.04 & \cellcolor{tabfirst}\textbf{0.01} & 
                               - & 
                               \cellcolor{tabfirst}\textbf{0.00} & \cellcolor{tabfirst}\textbf{0.00} & \cellcolor{tabfirst}\textbf{0.73} & \cellcolor{tabfirst}\textbf{0.11} & \cellcolor{tabfirst}\textbf{0.22} & \cellcolor{tabfirst}\textbf{0.89} 
                               \bigstrut[b]\\
\hline
\end{tabular}%
}
\vspace{-2mm}
\caption{Quantitative results on DigitalTwinArt-PM Dataset. We report averaged metrics over 10 trials with different random seeds. Joint 1 of ``Storage'' is prismatic, so there is no Axis Pos. The \colorbox{tabfirst}{best} and the \colorbox{tabsecond}{second best} results are highlighted.
}
\label{tab:3_part}
\end{table*}
\begin{table*}[t]
\centering
\def\mywidth{0.99\textwidth} 
\resizebox{\mywidth}{!}{
{\fontsize{3.9}{4}\selectfont
\begin{tabular}{lc |ccccccc}
\hline
                               &                                & \multicolumn{1}{c}{Axis Ang} & \multicolumn{1}{c}{Axis Pos} & \multicolumn{1}{c}{Part Motion} & \multicolumn{1}{c}{CD-s}       & \multicolumn{1}{c}{CD-m}     & \multicolumn{1}{c}{CD-w}\bigstrut\\
\hline
\multirow{2}{*}{\shortstack{Table - { 19836 (4 parts)}}} 
  & DigitalTwinArt \cite{weng2024digitaltwin} & 51.74 & - & 0.80 & 10.56 & 133.87 & \textbf{0.96}  \bigstrut[t]\\
  & ArtGS \cite{liu2025artgs} & 15.22 & - & 0.08 & 2.98 & 115.71 & 1.62  \bigstrut[t]\\
  & Ours 
  & \textbf{0.01} 
  & - 
  & \textbf{0.00}
  & \textbf{0.79}
  & \textbf{0.27} 
  & 1.50 \bigstrut[b]\\
\hline
\multirow{2}{*}{\shortstack{Table - { 25493 (4 parts)}}} 
  & DigitalTwinArt \cite{weng2024digitaltwin} & 25.30 & - & 0.39 & 1.08 & 308.51 & \textbf{0.61} \bigstrut[t]\\
  & ArtGS \cite{liu2025artgs} & 21.59 & - & 0.05 & 0.55 & 109.13 & 0.66  \bigstrut[t]\\
  & Ours 
  & \textbf{0.02} 
  & - 
  & \textbf{0.00} 
  & \textbf{0.41}
  & \textbf{0.09}
  & 0.67 \bigstrut[b]\\
\hline
\multirow{2}{*}{\shortstack{Storage - { 45503 (4 parts)}}} 
  & DigitalTwinArt \cite{weng2024digitaltwin} & 41.77 & 3.18 & 26.60 & 2.40 & 176.78 & 0.98 \bigstrut[t]\\
  & ArtGS \cite{liu2025artgs} & 4.86 & 2.82 & 10.28 & 1.29 & 3.20 & 4.86 \bigstrut[t]\\
  & Ours 
  & \textbf{0.02} 
  & \textbf{0.01} 
  & \textbf{0.09}
  & \textbf{0.70}
  & \textbf{0.10}
  & \textbf{0.97} 
  \bigstrut[b]\\
\hline
\multirow{2}{*}{\shortstack{Storage - { 41083 (5 parts)}}}
  & DigitalTwinArt \cite{weng2024digitaltwin} & 61.11 & 5.30 & 34.48 & 2.39 & 355.04 & \textbf{1.16} \bigstrut[t]\\
  & ArtGS \cite{liu2025artgs} & 1.08 & 1.65 & 10.28 & 1.98 & 120.15 & 2.05 \bigstrut[t]\\
  & Ours  
  & \textbf{0.03} 
  & \textbf{0.01} 
  & \textbf{0.06}
  & \textbf{0.94}
  & \textbf{0.16}
  & 6.77 \bigstrut[b]\\
\hline
\multirow{2}{*}{\shortstack{Storage - { 46145 (7 parts)}}} 

  & DigitalTwinArt \cite{weng2024digitaltwin} & 64.77 & 4.30 & 25.68 &1.58  & 400.71 & \textbf{0.84} \bigstrut[t]\\
  & ArtGS \cite{liu2025artgs} & 8.49 & 3.30 & 12.29 & 1.54 & 359.14  & 2.30 \bigstrut[t]\\
  & Ours 
  & \textbf{0.07} 
  & \textbf{0.00}
  & \textbf{0.06}
  & \textbf{0.52}
  & \textbf{0.09}
  & 1.58
\bigstrut[b]\\
\hline
\end{tabular}
}
}
\vspace{-2mm}
\caption{Quantitative results on GS-PM Dataset. Metrics are averaged over 5 trials with different random seeds; see supplementary for details. "Table-19836" and "Table-25493" has 3 prismatic joints with no Axis Pos.
}
\label{tab:multi_part}
\end{table*}

\begin{figure*}[h!]
    \centering
    \includegraphics[width=0.95\linewidth]{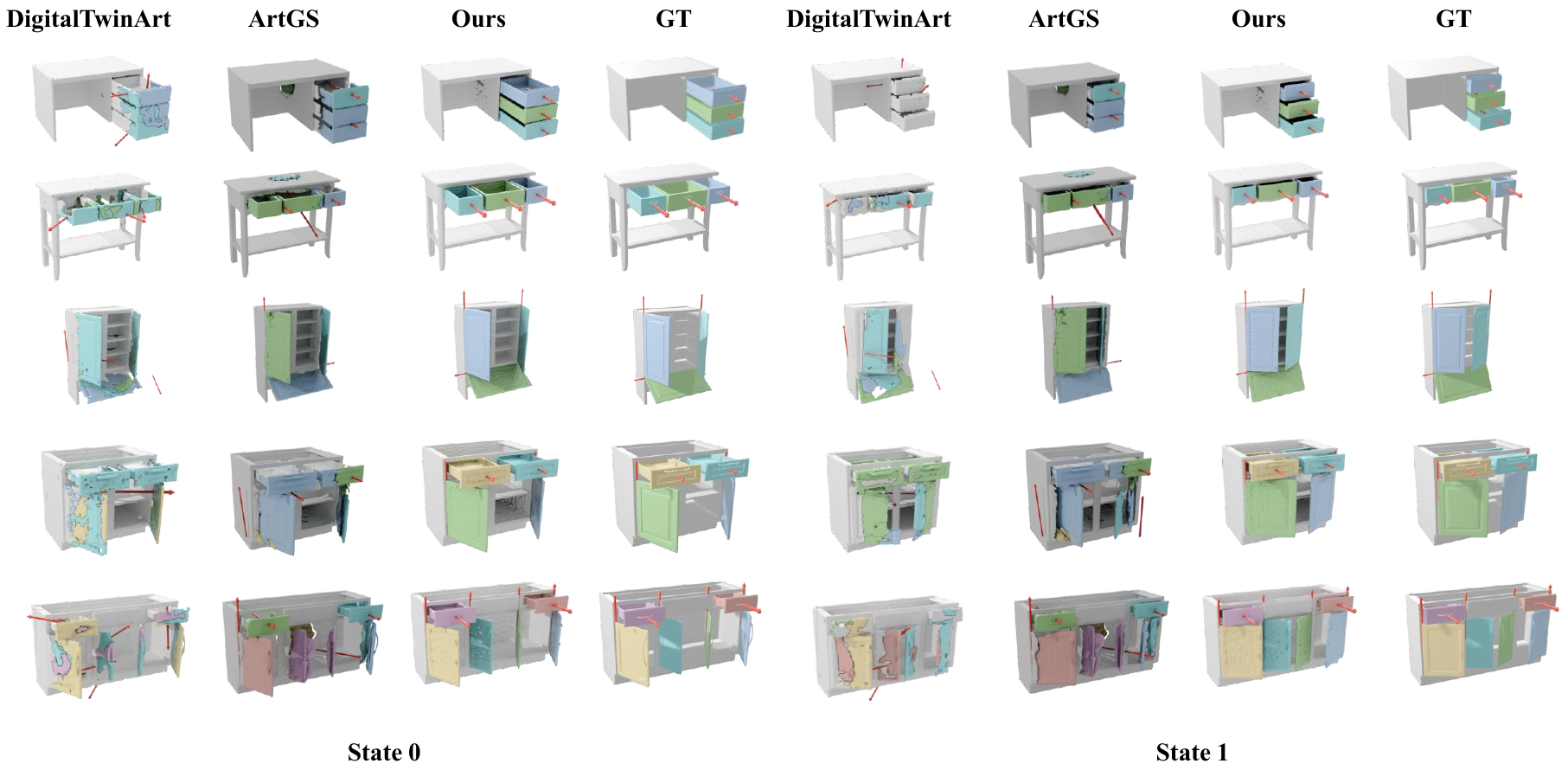}
    \vspace{-2mm}
    \caption{Qualitative results of multi-part objects. Top-to-bottom: Table-19836; Table-25493; Storage-45503; Storage-41083; Storage-46145.}
    \label{fig:mpm_supp}
    \vspace{-4mm}
\end{figure*}

\subsection{Experiments on PARIS Two-Part Dataset}
\noindent\textbf{Implementation Details.} Following \cite{weng2024digitaltwin},
we compare against Ditto \cite{Jiang_2022_ditto}, PARIS \cite{liu2023paris}, PARIS* (PARIS augumented with depth supervision),  CSG-reg \cite{weng2024digitaltwin}, 3Dseg-reg \cite{weng2024digitaltwin}, DigitalTwinArt \cite{weng2024digitaltwin}, and ArtGS \cite{liu2025artgs}. 
For a fair evaluation, we follow DigitalTwinArt and report the mean $\pm$ std for each metric over the 10 trials at the high-visibility state. 

\noindent\textbf{Results.}
As depicted in \cref{tab:two_part}, our GaussianArt significantly outperforms other baselines in PARIS Two-part Dataset in both motion parameter estimation and geometric reconstruction. For motion parameter estimation, especially on simulated data, GaussianArt and ArtGS achieve near-zero errors and substantially outperform DigitalTwinArt. Even on real-world data with noisy depth information, GaussianArt maintains high accuracy, thanks to its effective soft-to-hard optimization approach. In geometric reconstruction, GaussianArt excels over all competing methods for static and dynamic parts. Furthermore, GaussianArt achieves superior visual quality compared to other SoTAs (\cref{tab:2pr}), demonstrating strong potential for realistic digital twins of articulated objects.

\subsection{Experiments on Multi-Part Dataset}
\noindent\textbf{Implementation Details.} On the DigitalTwinArt‑PM Dataset, we compare against PARIS* (PARIS augmented with depth supervision), DigitalTwinArt\cite{weng2024digitaltwin}, and ArtGS\cite{liu2025artgs}. On the GS‑PM Dataset, we compare against DigitalTwinArt\cite{weng2024digitaltwin} and ArtGS\cite{liu2025artgs}.
We report the mean for each metric over the 5 trials at the high-visibility state.

\noindent\textbf{Results.}
As depicted in \cref{tab:3_part}, both GaussianArt and ArtGS demonstrate strong performance on objects with 3 parts, indicating that their GS-based motion field modeling frameworks are fundamentally sound. However, as shown in \cref{tab:multi_part}, when the number of object parts increases, ArtGS exhibits instability in both part segmentation and motion parameter learning. In contrast, GaussianArt, benefiting from a strong prior for initialization, is able to model multi-part objects in a stable manner.

\cref{fig:mpm_supp} further illustrates that ArtGS heavily relies on the initialization of clusters for multi-part objects. In some cases, even manual adjustments to the cluster positions fail to yield clear part segmentation, resulting in inaccurate motion parameter predictions. GaussianArt, on the other hand, leverages a vision foundation model to perform stable pre-segmentation of the object, enabling accurate motion learning and offering better scalability to larger benchmarks in the future.

\section{Additional Qualitative Results}
\label{sec:17}
\begin{figure*}[t] 
\centerline{
\includegraphics[width=0.95\linewidth]{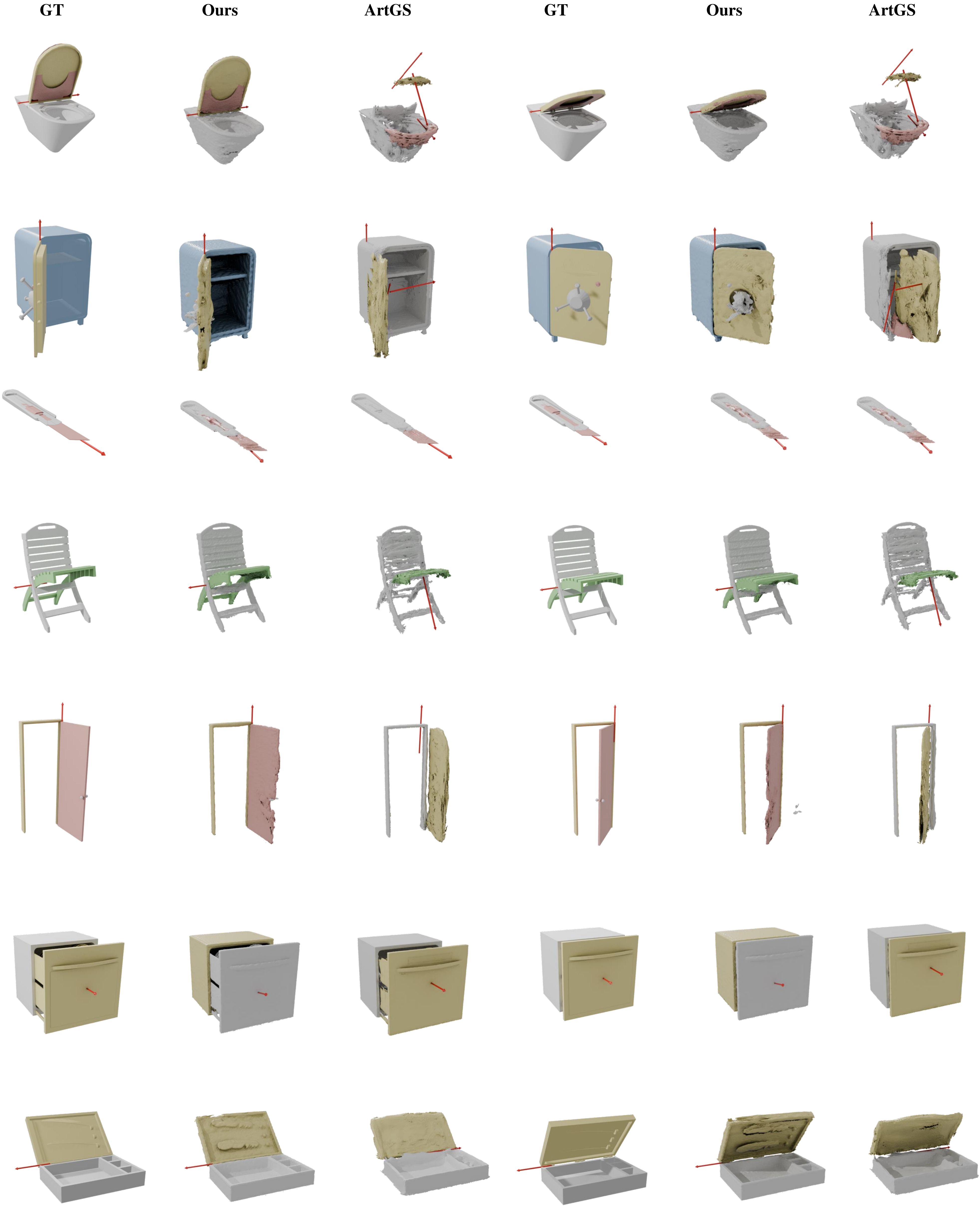}}
\vspace{-2mm}
\caption{Other qualitative results on ReconArticulate.}
\label{Fig:visual2} 
\end{figure*}

\cref{Fig:visual2} shows additional qualitative comparisons on MPArt-90.

\section{MPArt-90}
\label{sec:18}
\begin{figure*}[t] 
\centerline{
\includegraphics[width=0.95\linewidth]{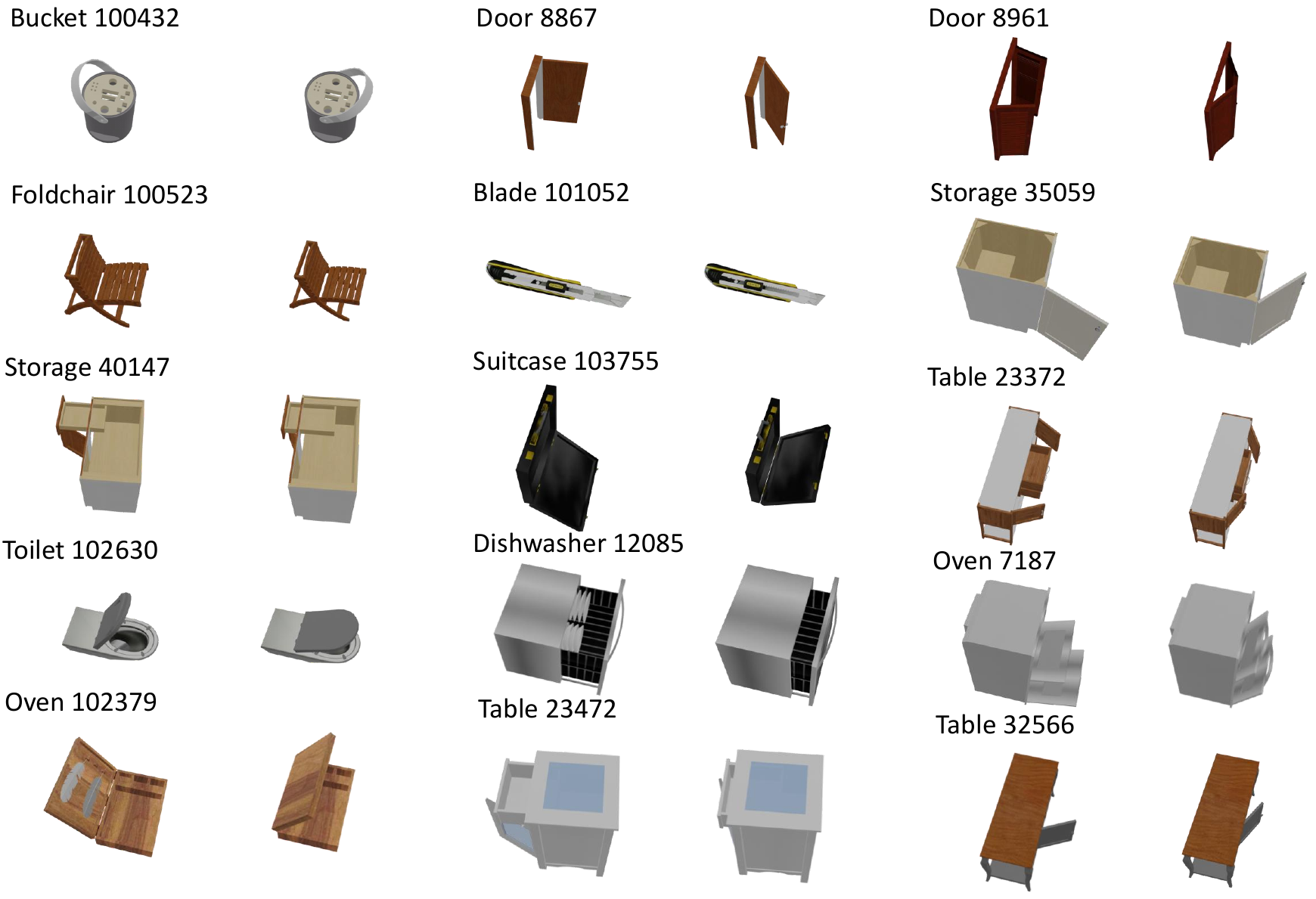}}
\vspace{-2mm}
\caption{Some examples of MPArt-90.}
\label{Fig:example} 
\end{figure*}

\cref{tab:MPArt-90-st} shows the statistics of our MPArt-90 benchmark. It contains 90 objects from 20 categories, including 12 objects from PARIS Two-Part Dataset \cite{liu2023paris} and 2 objects from DigitalTwinArt-PM Dataset \cite{weng2024digitaltwin}. \cref{Fig:example} shows some examples of MPArt-90.

\begin{table*}[htbp]
\centering
\begin{tabular}{l|ccccccccccc|c}
\toprule
\diagbox{Category}{Part number} & 2 & 3 & 4 & 5 & 6 & 7 & 8 & 9 & 10 & 15 & 20 & Overall \\
\midrule
Storage     & 9 & 9 & 7 & 5 & -- & 3 & -- & 1 & -- & 1 & 1 & 36 \\
Table       & 5 & 7 & 5 & 4 & 1 & -- & 1 & 2 & 1 & -- & -- & 26 \\
Blade       & 2 & -- & -- & -- & -- & -- & -- & -- & -- & -- & -- & 3 \\
Fridge      & 2 & 1 & -- & -- & -- & -- & -- & -- & -- & -- & -- & 3 \\
Oven        & 1 & 2 & -- & -- & -- & -- & -- & -- & -- & -- & -- & 3 \\
Window      & -- & -- & 1 & -- & -- & -- & -- & -- & -- & -- & -- & 1 \\
Washer      & 2 & -- & -- & -- & -- & -- & -- & -- & -- & -- & -- & 2 \\
Foldchair   & 2 & -- & -- & -- & -- & -- & -- & -- & -- & -- & -- & 2 \\
Door        & -- & 2 & -- & -- & -- & -- & -- & -- & -- & -- & -- & 2 \\
Bucket      & 1 & -- & -- & -- & -- & -- & -- & -- & -- & -- & -- & 1 \\
Suitcase    & 1 & -- & -- & -- & -- & -- & -- & -- & -- & -- & -- & 1 \\
Box         & 1 & -- & -- & -- & -- & -- & -- & -- & -- & -- & -- & 1 \\
DishWasher  & 1 & -- & -- & -- & -- & -- & -- & -- & -- & -- & -- & 1 \\
Safe        & 1 & -- & -- & -- & -- & -- & -- & -- & -- & -- & -- & 1 \\
Laptop      & 1 & -- & -- & -- & -- & -- & -- & -- & -- & -- & -- & 1 \\
Toilet      & 1 & -- & -- & -- & -- & -- & -- & -- & -- & -- & -- & 1 \\
USB         & 1 & -- & -- & -- & -- & -- & -- & -- & -- & -- & -- & 1 \\
Scissor     & 1 & -- & -- & -- & -- & -- & -- & -- & -- & -- & -- & 1 \\
Stapler     & 1 & -- & -- & -- & -- & -- & -- & -- & -- & -- & -- & 1 \\
Microwave   & 1 & -- & -- & -- & -- & -- & -- & -- & -- & -- & -- & 1 \\
\midrule
Overall     & 34 & 21 & 14 & 10 & 1 & 3 & 1 & 3 & 1 & 1 & 1 & 90 \\
\bottomrule
\end{tabular}
\caption{The statistics of MPArt-90.}
\label{tab:MPArt-90-st}
\end{table*}

\end{document}